\documentclass{article}

\usepackage{hyperref}
\usepackage{url}

\usepackage{thm-restate}

\usepackage{comment}
\usepackage{amsmath}
\usepackage{amsthm}
\usepackage{amssymb}
\usepackage{algorithm}
\usepackage{algpseudocode}

\usepackage{subfig}
\usepackage[pdftex]{graphicx}
\usepackage{bm}
\usepackage{booktabs}
\usepackage{siunitx}
\usepackage{bbm}
\usepackage{wrapfig}

\newcommand{\oo}{\mathcal{O}}
\newcommand{\E}{\mathbb{E}}
\newcommand*{\op}{\mathsf{op}}

\def\gN{{\mathcal{N}}}

\def\gV{{\mathcal{V}}}

\usepackage{comment}
\newtheorem{theorem}{Theorem}[section]
\newtheorem{lemma}[theorem]{Lemma}

\newtheorem{definition}[theorem]{Definition}

\newtheorem{corollary}[theorem]{Corollary}

     \usepackage[preprint]{neurips_2024}

\usepackage[utf8]{inputenc} % allow utf-8 input
\usepackage[T1]{fontenc}    % use 8-bit T1 fonts
\usepackage{hyperref}       % hyperlinks
\usepackage{url}            % simple URL typesetting
\usepackage{booktabs}       % professional-quality tables
\usepackage{amsfonts}       % blackboard math symbols
\usepackage{nicefrac}       % compact symbols for 1/2, etc.
\usepackage{microtype}      % microtypography
\usepackage{xcolor}         % colors

\title{Flood and Echo Net:\\ Algorithmically Aligned GNNs that Generalize} 

\author{%
Joël Mathys \\
  ETH Zurich\\
  \texttt{jmathys@ethz.ch} \\
   \And
   Florian Grötschla \\
   ETH Zurich \\
  \texttt{fgroetschla@ethz.ch} \\
     \And
   Kalyan Varma Nadimpalli \\
   IIIT Bangalore \\
  \texttt{kalyan.varma@iiitb.ac.in} \\
     \And
   Roger Wattenhofer \\
   ETH Zurich \\
  \texttt{wattenhofer@ethz.ch} \\
}

\begin{document}

\maketitle

\begin{abstract}
Most Graph Neural Networks follow the standard message-passing framework where, in each step, all nodes simultaneously communicate with each other. 
We want to challenge this paradigm by aligning the computation more closely to the execution of distributed algorithms and propose the Flood and Echo Net. 
A single round of a Flood and Echo Net consists of an origin node and a flooding phase followed by an echo phase. First, during the flooding, messages are sent from the origin and propagated outwards throughout the entire graph. Then, during the echo, the message flow reverses and messages are sent back towards the origin. 
As nodes are only sparsely activated upon receiving a message, this leads to a wave-like activation pattern that traverses the graph. Through these sparse but parallel activations, the Net becomes more expressive than traditional MPNNs which are limited by the 1-WL test and also is provably more efficient in terms of message complexity.
Moreover, the mechanism's inherent ability to generalize across graphs of varying sizes positions it as a practical architecture for the task of algorithmic learning. We test the Flood and Echo Net on a variety of synthetic tasks and the SALSA-CLRS benchmark and find that the algorithmic alignment of the execution improves generalization to larger graph sizes. 
\end{abstract}

%===================
% Introduction
%===================
\section{Introduction}

The message-passing paradigm lies at the very center of graph learning. Many of the proposed methods and approaches belong to the framework of Message Passing Neural Networks (MPNNs). In these networks, every node maintains a state and updates it based on the states of all its neighbors by exchanging messages after every round. Through this operation principle, MPNNs can provide the necessary flexibility to be applied to arbitrary graph topologies. However, this simple procedure needs a considerable amount of compute, as in every round, messages have to be sent over all edges. Moreover, all nodes throughout the entire graph update their state. Even though it might be that the majority of the nodes do not play an active part in the computation and maintain their current state, resulting in unnecessary computations. This phenomenon is potentially amplified if the network is applied to graphs of larger sizes, as the number of rounds might have to be scaled in order to capture non local interactions between nodes. We challenge this paradigm and instead base the mechanism of Graph Neural Networks on another principled way, inspired by an algorithm design pattern commonly found in distributed computing. 

\begin{figure}
  \centering
    \includegraphics[width=\linewidth]{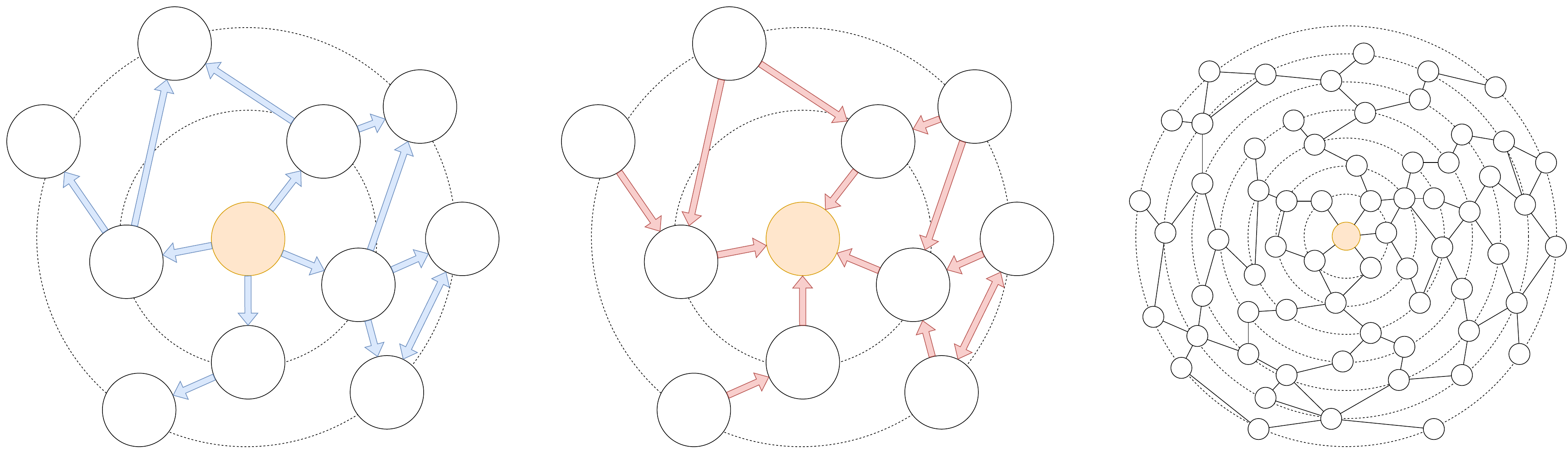}
    \caption{The Flood and Echo Net propagates messages in a wave-like pattern throughout the entire graph. Starting from an origin (orange), messages are sent toward the origin's neighbors and then continuously sent or ``flooded'' farther away outwards (blue). Afterward, the flow reverses, and messages are ``echoed'' back (red) toward the origin. Throughout the computation, only a small subset of nodes is active at any given time, passing messages efficiently throughout the entire graph. Moreover, the mechanism naturally generalizes to graphs of larger sizes.}
    \label{fig:intro}
\end{figure}

We therefore propose a new execution framework, the Flood and Echo Net. Although it is based on message-passing, the strategy by which messages are exchanged is distinct from regular MPNNs. Instead, the execution aligns with an algorithmic design pattern commonly used in distributed computing called \textit{flooding and echo}. The computation is initiated by a single node, called the origin, which initiates a phase consisting of a flooding and an echo part, respectively. First, messages are propagated away from the root towards the rest of the graph. In this flooding part nodes only send messages to nodes that are farther away from the origin. Once all nodes have received a message, the propagation flow reverses. Now, nodes only send messages to neighbors closer to the origin, starting with the nodes that are farthest away. An illustration of this is shown in Figure \ref{fig:intro}. This activation pattern lies at the very center of the Flood and Echo Net and can be thought of as a wave pattern that equally extends outwards in all directions and then reverses back to the origin. 

Note that during the execution of Flood and Echo Net, only a subset of nodes is active at every single timestep. This set consists of nodes that are located at the same distance and send messages either further away or back towards the origin. This results in a sparse but, at the same time, very parallel activation pattern throughout the graph. Moreover, the computation involves the entire graph while using relatively few messages. We find that compared to regular MPNNs, the Flood and Echo Net differs in three important aspects that are of interest for GNNs, which include message complexity, expressivity, and the way the mechanism generalizes to larger graph sizes. 

Standard MPNNs exchange information with their one-hop neighborhood in each round, sending $\oo(m)$ messages in total along all edges. On the other hand, an entire phase of a Flood and Echo Net also only exchanges $\oo(m)$ messages. However, during this process, the update of a node's state can take into account information beyond its immediate local neighborhood. This can ultimately result in improved message complexity. Moreover, as the computation implicitly leverages distance information in the graph, the expressiveness of the Flood and Echo Net goes beyond the 1-WL test.  
Finally, we turn to the topic of algorithm learning, where GNNs learn algorithms from data which should then generalize across graph sizes. When MPNNs are applied to graphs of larger sizes, they must usually adapt and scale the number of rounds according to the graph size to retain the same relative field of perception. In comparison, the execution of the Flood and Echo Net can generalize to graphs of larger sizes more naturally as the computation inherently involves the entire graph. We hypothesize that through the algorithmic alignment of the underlying mechanism, the Flood and Echo Net is of interest for algorithm learning. We test our architecture on a variety of algorithmic tasks, including the SALSA-CLRS benchmark, and find that the Flood and Echo Net can improve the ability to generalize to larger graphs. 

We outline our main contributions as follows: 
\begin{itemize}
    \item We introduce the Flood and Echo Net, a new execution framework aligned with principles of distributed algorithm design. The computation follows a special node activation pattern, which allows it to facilitate messages more efficiently throughout the graph. 
    \item We prove that through this alternative computation flow, the Flood and Echo Net is more expressive than common MPNNs and more efficient in terms of message complexity. 
    \item We show that the algorithmic alignment of the architecture is beneficial for size generalization in algorithm learning. We empirically validate our findings on a variety of synthetic tasks and test on the SALSA-CLRS benchmark. 
\end{itemize}

%===================
% Related Work
%===================
\section{Related Work}

Originally proposed by \citet{scarselli2008graph}, Graph Neural Networks have seen a resurgence with applications across multiple domains \citep{gat_2017, kipf2016semi,pmlr-v220-neun22a}. Notably, this line of research has gained theoretical insights through its connection to message-passing models from distributed computing \citep{sato2019approximation, loukas2020graph, papp2022distr}. This includes strengthening existing architectures to achieve maximum expressiveness \citep{xu2018powerful, sato2021random} or going beyond traditional models by changing the graph topology \citep{papp2021dropgnn, alon2021on}.
In this context, multiple architectures have been investigated to combat information bottlenecks in the graph \citep{alon2021bottleneck}, i.e. using graph transformers \citep{rampasek2022recipe}. Note that our work is orthogonal to this, as we focus on message-passing on the original graph topology. Moreover, we investigate how specific information can be exchanged throughout the entire graph, which might be challenging even if no bottleneck is present. Similarly, higher order propagation mechanisms \citep{zhang2023rethinking, maron2020provably, zhao2022practical} have been proposed to tackle this issue or gain more expressiveness. 
While some of these approaches also incorporate distance information, this usually comes at the cost of higher-order message-passing. Whereas our work emphasizes a simple execution mechanism on the original graph topology. 
In recent work, even the synchronous message-passing among all nodes has been questioned \citep{martinkus2023agentbased, faber2023asynchronous}, giving rise to alternative neural graph execution models. 

How GNNs can generalize across graph sizes \citep{yehudai2021local} and especially their generalization capabilities for algorithmic tasks, attributed to their structurally aligned  computation \citep{xu2020neural} has been of much interest. This has led to investigations into the proper alignment of parts of the architecture \citep{dudzik2022graph, engelmayer2023parallel, dudzik2023asynchronous}. A central focus has been on how these networks learn to solve algorithms \citep{pmlr-v162-velickovic22a, ibarz2022generalist, minder2023salsaclrs}. Moreover, the ability to extrapolate \citep{xu2021neural} and dynamically adjust the computation in order to reason for longer when confronted with more challenging instances remains a key aspect \citep{NEURIPS2021_3501672e, grötschla2022learning, tang2020towards}.

%===================
% Flood and Echo Net
%===================
\section{Flood and Echo Net}

The fields of distributed computing and Graph Neural Networks are interlinked through their shared usage of message-passing-based computation. While there are differences, the connection and equivalence between certain models were established \citep{papp2022distr}. Through this connection, several results, such as theoretical bounds regarding width, number of rounds, and approximation ratios, could be translated directly to GNNs \citep{sato2019approximation, loukas2020graph}. Moreover, it was shown that aligning the architecture well with the underlying learning objective \citep{xu2020neural, dudzik2022graph} can be beneficial both in terms of performance and sample complexity. This begs the question of whether we could potentially transfer other insights from distributed computing to the field of graph learning. 
Therefore, we propose the Flood and Echo Net, an alternative execution framework that aligns with design patterns from distributed algorithms.

First, recall the standard execution of a message-passing-based GNN. 
Whenever we refer to an MPNN throughout this paper, we will refer to a GNN that operates on the original graph topology and exchanges messages in the following way: 
\begin{align*}
    a_v^{t} &= \text{AGGREGATE}^k(\{\!\{x_u^t \mid u \in N(v)\}\!\})\\
    x_v^{t+1} &= \text{UPDATE}(x_v^t, a_v^t)
\end{align*}
Note that all nodes exchange messages with all their neighbors in every round. We challenge this paradigm by taking inspiration from a design pattern called \textit{flooding and echo} \citep{changEcho}. This pattern is a common building block in distributed algorithms \citep{Kuhn2007tight} to first broadcast (flooding) \citep{Dalal1978ReversePF} messages throughout the entire graph and then gather back (echo) information from all nodes. 

\begin{figure}[ht]
    \centering
    \begin{minipage}[t]{0.48\textwidth}
        \vspace{-196pt} % Adjust this value as needed
        \begin{algorithm}[H]
            \caption{Flood and Echo Net}
            \begin{algorithmic}[1]
            
                        \State $D \gets \text{distances}(G, \text{origin})$
        \State $\text{maxD} \gets \text{max}(D)$
        \State $x \gets \text{Encoder}(x)$
        \For{$\text{t} = 1$ to $\text{phases}$}
            \For{$\text{d} = 1$ to $\text{maxD}$} flooding
                \State $x[d] \gets \text{FConv}^t(d-1\to d)$
                \State $x[d] \gets \text{FCrossConv}^t({d}\to {d})$
            \EndFor
            \For{$\text{d} = \text{maxD}$ to $1$} echo
                \State $x[d] \gets \text{ECrossConv}^t(d\to d)$
                \State $x[d-1] \gets \text{EConv}^t(d\to d-1)$
            \EndFor
            \State $x \gets \text{Update(x)}$
        \EndFor
        \State $x \gets \text{Decoder}(x)$
            \end{algorithmic}
        \end{algorithm}
    \end{minipage}%
    \hfill
    \begin{minipage}[t]{0.45\textwidth}
        \centering
        \includegraphics[width=\linewidth]{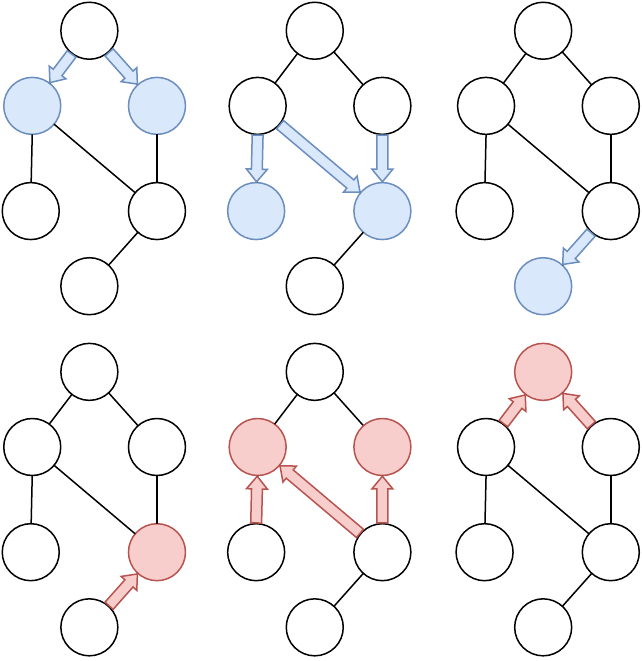} % Path to your image
        %\captionof{figure}{Sample Figure}
    \end{minipage}
    \caption{On the left, an algorithm describing the Flood and Echo Net. First, the distances are pre-computed to activate and update the proper nodes. The convolutions $a \to b$ send messages from nodes at distance $a$ to nodes at distance $b$, with only the nodes at distance $b$ updating their state, indicated by $x[b]$. On the right, an illustration of a single phase of a Flood and Echo Net. At every update step, only a subset of nodes is active and changes its state. The origin is the top node of the graph, and the blue arrows depict the information flow in the flooding, while the red arrows represent the echo part. Note that during a single phase activates all nodes in the graph, regardless of the graph size, while sending only a constant number of messages across each edge. }
    \label{fig:flood_and_echo}
\end{figure}

We intend to align the underlying computation flow directly with this design pattern. In the Flood and Echo Net, the computation is initiated from an origin node. Then, $T$ phases, each consisting of a flooding and echo part, are executed. In Figure \ref{fig:flood_and_echo} we outline the pseudo code for the Flood and Echo Net. At the beginning, nodes are partitioned according to their distance to the origin. Then, $T$ phases are executed, in each phase a flooding followed by an echo is performed. During the flooding part, messages propagate outwards, away from the origin. To achieve this, we iterate through the distances in ascending order. We differentiate between two types of edges in the convolutions. First, FConv sends messages from nodes at distance $d-1$ towards nodes at distance $d$. However, only the nodes at distance $d$ update their state, indicated by the notation $x[d]$. Then, FCrossConv sends messages between nodes that are at distance $d$. 
After the completion of the flooding part, the message flow reverses and is echoed back towards the origin. Again, we iterate over the distances but now in descending order. Similarly to above, we distinguish updates of nodes at the same distance using ECrossConv and updating nodes at distance $d-1$, which receive messages from nodes at distance $d$ through EConv.
Note that usually, only a subset of nodes, which are located at the same distance, are activated simultaneously. Therefore, Flood and Echo Net can make use of a sparse but parallel activation pattern that propagates throughout the entire graph. For a visual illustration of an entire phase, we refer to Figure \ref{fig:flood_and_echo}. The colors indicate which edges are used and which nodes are updated. For a more in-depth discussion of the definition, we refer to Appendix \ref{app:flood_and_echo}, which also contains a comparison to the computation of regular MPNNs. 

\paragraph{Modes of Operation} The computation of the Flood and Echo Net starts from an origin node. This allows for different usages of the proposed method. In the following, we outline three different strategies, which we will refer to as different modes of operations: \textit{fixed, random} and \textit{all}. 
Across all modes of operation, once the origin is chosen, the same flooding and echo parts are executed to compute node embeddings. These are directly used for node classification tasks; for graph classification, we sum up the final predicted class probabilities of the individual nodes. 

In the \textit{fixed} mode, the origin is given or defined by the problem instance, i.e. by a marked source node specific to the task. Alternatively, in the \textit{random} mode, an origin is chosen amongst all nodes uniformly at random. In the \textit{all} mode, we execute the Flood and Echo Net once for every node. In every run, we keep only the node embedding for the chosen origin. This can be seen as a form of ego graph prediction \cite{zhao2021gophormer} for each node. Although computationally more expensive, it could also be used for efficient inference on tasks where only a subset of nodes is of interest.

%===================
% Theory
%===================
\section{Theoretical Analysis}

Throughout this section, we analyze the properties of the Flood and Echo Net. While the presented execution framework is still based on message-passing on the original graph topology, its propagation strategy distinguishes it from regular message-passing GNNs. 
Specifically, we show that even though the mechanism remains simple and only exchanges messages on the original graph topology, it is more expressive than regular MPNNs. Moreover, due to the sparse activation of nodes and the order of how messages are exchanged, the Flood and Echo Net is more efficient regarding message complexity. As a consequence, there exist tasks that can be solved by a Flood and Echo Net using $\oo(m)$ messages, which cannot be solved by MPNNs without using asymptotically more messages. For the complete proofs of the following theorems, we refer to Appendix \ref{app:proofs}.

\subsection{Expressiveness}
\begin{wrapfigure}{r}{0.5\textwidth}
    \centering
\includegraphics[width=\linewidth]{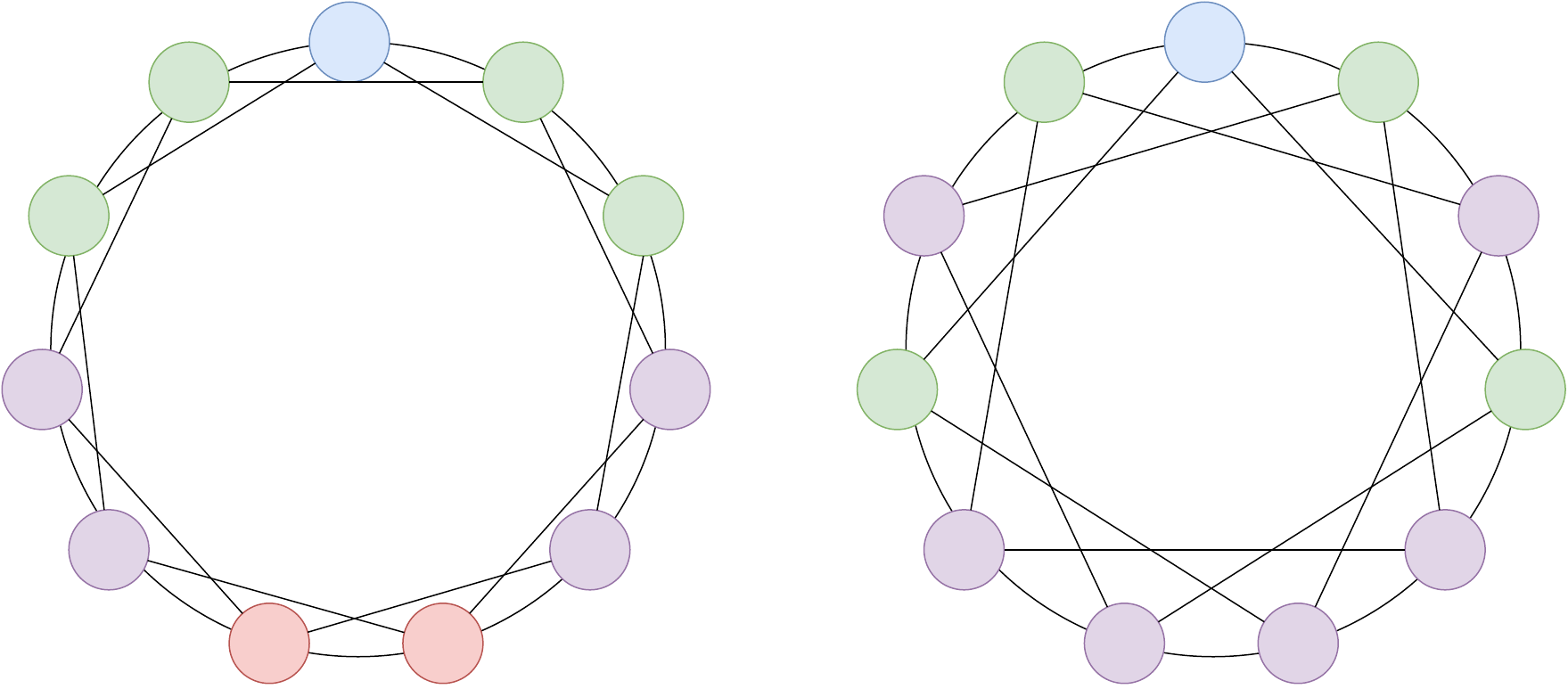}
\caption{Example of two $4$-regular graphs which cannot be distinguished using standard MPNNs as they are 1-WL equivalent. However, no matter which starting point is chosen, the Flood and Echo Net can easily distinguish between them through the derived distance to the starting node.}
\label{fig:expressive}
%\vspace{-1cm}
\end{wrapfigure}
The expressiveness of GNNs is tightly linked to the WL test \citep{wl_1968}. Most common message-passing architectures, which work on the original graph topology without higher-order message-passing, are limited by the expressiveness of 1-WL \citep{papp2022theoretical}. 
First, we show that even though the Flood and Echo Net operates differently from MPNNs, it can still match them in their expressiveness and, therefore, remain maximally expressive in terms of the 1-WL test. Moreover, it can do so using at most the same number of messages as a regular MPNN.

\begin{theorem}
\label{thm:flood_echo_expressiveness} On connected graphs, 
    the Flood and Echo Net is at least as expressive as any MPNN . Furthermore, it exchanges at most as many messages. 
\end{theorem}

However, while MPNNs are limited by the 1-WL test, the Flood and Echo Net is more expressive. Although it also exchanges messages solely on the original graph topology, the mechanism can implicitly leverage more information to distinguish more nodes through the alignment of the message propagation with the distance to the origin in the graph.  

\begin{theorem}
\label{thm:flood_echo_wl} On connected graphs,
    Flood and Echo Net is strictly more expressive than 1-WL and, by extension, standard MPNNs. 
\end{theorem}

To get an intuition on this insight, we can think about how the flooding and echo mechanism differs from the perspective of a single node. Usually, all edges send and receive messages in every round. Therefore, the edges are identical. In the Flood and Echo Net, we introduce a ``direction'' of the edges, and we can distinguish between edges that go towards or away from the origin or if they have the same distance. This gives us more possibilities to distinguish nodes in our local neighborhood and leverage non-local information as the wave pattern transitions through the whole graph. At the same time, the net could ignore this additional directionality information of the edges and simulate the execution of a standard MPNN. Next to these theoretical insights, we also empirically validate that the Flood and Echo Net is more expressive on a variety of datasets which we include in Appendix \ref{app:expressive_results}.

\subsection{Message Complexity}
In regular MPNNs, a single round of message-passing updates all $n$ node states by exchanging messages over all edges. Therefore, every single round exchanges $\oo(m)$ messages while propagating information by one hop. If any information needs to be propagated over a distance of $D$ hops, the total number of node updates is $\oo(Dn)$ and the total number of exchanged messages is $\oo(Dm)$.  

On the other hand, during the execution of a Flood and Echo Net, only a subset of nodes is active during each timestep, which sends messages away or towards the origin. This results in the key difference that nodes are sequentially activated, and the messages pass along information throughout the entire graph instead of only their immediate one-hop neighborhood. More precisely, in a single phase of a Flood and Echo Net, which consists of one flooding followed by one echo part, each node is activated a constant number of times, while there are also at most a constant number of messages passed along each edge. Therefore, a single phase performs $\oo(n)$ node updates and exchanges $\oo(m)$ messages. Crucially, if information needs to be exchanged over a distance of $D$ hops, this can be achieved with a constant number of phases, as each phase exchanges information throughout the entire graph. Therefore, it is possible to exchange information over a distance of $D$ hops using only $\oo(m)$ messages compared to $\oo(Dm)$ messages used by MPNNs. 

\begin{lemma}
\label{lemma:flood_echo_messages}
    There exist tasks that Flood Echo can solve using $\oo(m)$ messages, whereas no MPNN can solve them using less than $\oo(nm)$ messages.
\end{lemma}

As a consequence of this insight, it follows that there must exist tasks that can be solved more efficiently using the Flood and Echo Net. If information must be exchanged throughout the entire graph, it can be that MPNNs must use $\oo(nm)$ messages, while a constant amount of Flood and Echo phases with $\oo(m)$ messages would suffice. Moreover, recall that by Theorem \ref{thm:flood_echo_expressiveness}, the Flood and Echo Net can simulate the execution of other MPNNs while using, asymptotically speaking, at most the same number of messages. For a more detailed discussion on the runtime and message complexity, we refer to Appendix \ref{app:runtime}.

%===================
% Extrapolation
%===================
\section{Generalization in Algorithmic Tasks}

In this section, we study algorithm learning using Graph Neural Networks. The concept of an algorithm is best understood as a sequence of instructions that can be applied to compute a desired output given the respective input. Algorithms have the inherent advantage of generalization across their entire domain. If we want to multiply two numbers, we can easily illustrate and explain the multiplication algorithm using small numbers. However, the same procedure generalizes, i.e. the algorithm can be used to extrapolate and multiply together much larger numbers using the same
algorithmic steps. Algorithm learning aims to grasp these underlying principles and incorporate them into machine learning architectures. The ultimate aim is to combine both domains to get models that can learn the algorithmic principles and generalize them properly even for unseen larger inputs. 

One difficulty when studying generalization is how the architecture should adapt to larger sizes. If it does not adjust at all, it might be that the amount of compute does not suffice to solve the task at hand, or in the case of graph tasks, that the required information is no longer located in the same receptive field but is farther away. Therefore, a common strategy is to adjust the compute, or number of rounds, according to the increase of the problem size. Recall that the Flood and Echo Net generalizes to larger sizes in a different way compared to regular MPNNs. In fact, during a single phase, messages propagate throughout the entire graph and can, therefore, be updated using information beyond the immediate neighborhood. Previous work has already indicated that changes in the architecture or so-called ``algorithmic alignment'' \citep{engelmayer2023parallel, dudzik2022graph, xu2020neural} can be beneficial for learning and generalization. In our work, we base the underlying mechanism on the flooding and echo paradigm, an algorithm design pattern from distributed computing. Therefore, we believe that the Flood and Echo Net can improve generalization for algorithm learning.

In the following, we empirically validate our hypothesis on a variety of tasks related to algorithm learning. First, we test the architecture on three synthetic algorithmic tasks, which allow us both fine-grained control and theoretical insights into what is needed to solve the tasks at hand. Then, we proceed to test on the more challenging SALSA-CLRS benchmark, which consists of well-known graph algorithms.

\subsection{Algorithmic Tasks}

In this part, we focus on three algorithmic tasks PrefixSum, Distance and Path Finding adapted to the graph domain in \citet{grötschla2022learning}. In the Distance task, nodes have to infer their distance to a marked node modulo $2$. For the Path Finding task, nodes in a tree have to predict whether they are part of the path between two given nodes. Finally, in the PrefixSum task, the cumulative sum modulo 2 has to be computed on a path graph. For a more detailed description of the datasets, we refer to Appendix \ref{appendix:algorithmic_datasets}. While these tasks seem to be more naive compared to real algorithms, their simplicity allows us to inspect and reason more precisely what is required to solve these tasks. For a more thorough analysis of the Flood and Echo Net on the PrefixSum task, we refer to Appendix \ref{app:inf_prop}.
\begin{corollary}
\label{cor:distance_od}
    Let $D$ be the diameter of the graph. In order to correctly solve the Distance, Path-finding, and PrefixSum tasks, nodes require information that is $\oo$(D) hops away. 
\end{corollary}

\begin{figure}[t]
  \centering
  \includegraphics[width=\linewidth]{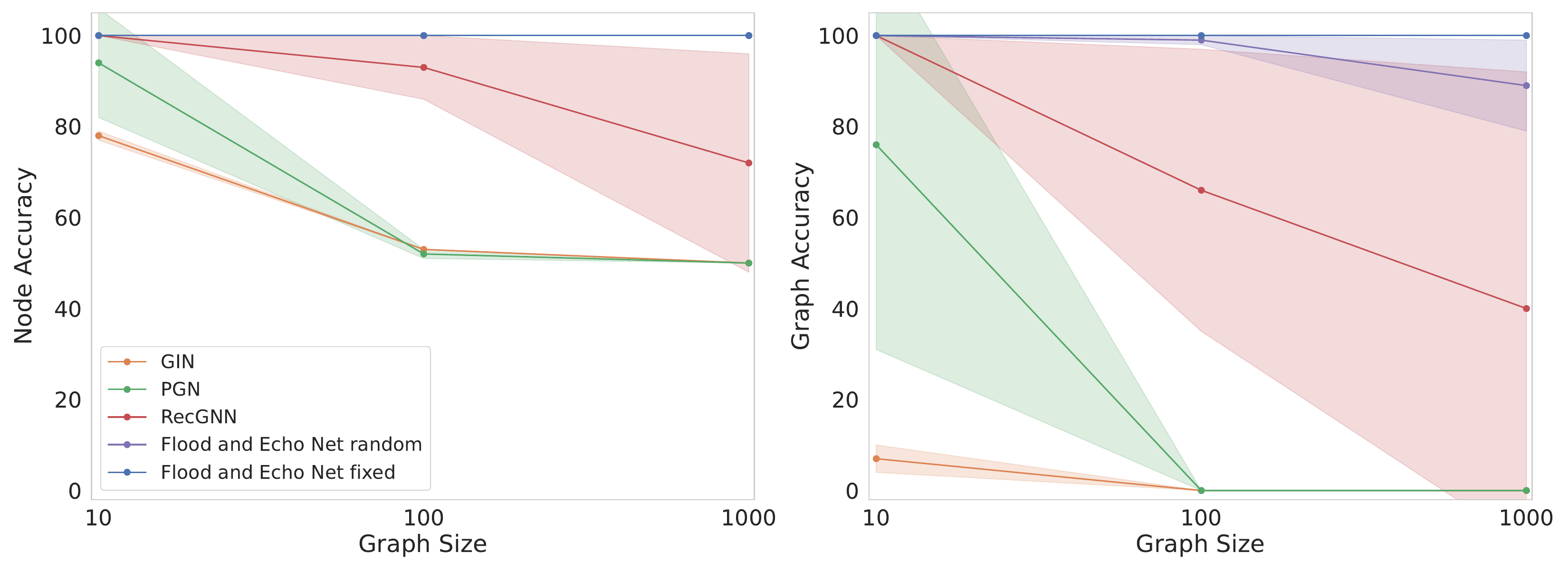}
  \caption{Extrapolation on the PrefixSum task. All models are trained with graphs of size 10 and then tested on larger graphs. The Flood and Echo models are able to generalize well to graphs $100$ times the sizes encountered during training. We report both the node accuracy on the left and the graph accuracy on the right. }
  \label{fig:prefix-plot}
\end{figure}

\begin{table*}[t]
\centering
\caption{Extrapolation experiments on all algorithmic datasets, all models were trained with graphs of size 10 and then tested on larger graphs of size $100$. We compare the different Flood and Echo models against a regular GIN, which executes $L$ rounds, PGN and RecGNN, which adapts the number of rounds. The \textit{random} mode picks a starting node at random, while the \textit{fixed} mode starts at a predefined location. The \textit{all} chooses each node as a start once. We report both the node accuracy with $n()$ and the graph accuracy with $g()$.}
\resizebox{1.0\textwidth}{!}{
\begin{tabular}{@{}l*{10}{S[table-format=-3.4]}@{}}
\toprule
{Model} & {\textsc{Messages}} & \multicolumn{3}{c}{\textsc{PrefixSum}} & \multicolumn{3}{c}{\textsc{Distance}} & \multicolumn{3}{c}{\textsc{Path Finding}} \\
{} & {} & {n(10)} & {n(100)} & {g(100)} & {n(10)} & {n(100)} & {g(100)} & {n(10)} & {n(100)} & {g(100)}\\
\midrule  
GIN & {$\oo(Lm)$}&{0.78 $\pm$ 0.01} & {0.53 $\pm$ 0.00 } & {0.00 $\pm$ 0.00 }&{0.97 $\pm$ 0.01} & {0.91 $\pm$ 0.01 } & {0.04 $\pm$ 0.06 }&{0.99 $\pm$ 0.01} & {0.70 $\pm$ 0.05 } & {0.00 $\pm$ 0.00 }\\
PGN & {$\oo(nm)$} &{0.94 $\pm$ 0.12} & {0.52 $\pm$ 0.01 } & {0.00 $\pm$ 0.00 }&{0.99 $\pm$ 0.01} & {0.89 $\pm$ 0.01 } & {0.01 $\pm$ 0.02 }&{1.00 $\pm$ 0.00} & {0.77 $\pm$ 0.03 } & {0.00 $\pm$ 0.00 }\\
RecGNN &  {$\oo(nm)$}&{1.00 $\pm$ 0.00} & {0.93 $\pm$ 0.07 } & {0.66 $\pm$ 0.31 }&{1.00 $\pm$ 0.00} & {0.99 $\pm$ 0.02 } & {0.93 $\pm$ 0.15 }&{1.00 $\pm$ 0.00} & {0.95 $\pm$ 0.04 } & {0.45 $\pm$ 0.33 }\\
\midrule
Flood and Echo \textit{all}& {$\oo(nm)$} &{1.00 $\pm$ 0.00} & {1.00 $\pm$ 0.01 } & {0.96 $\pm$ 0.07 }&{1.00 $\pm$ 0.00} & {0.99 $\pm$ 0.03 } & {0.87 $\pm$ 0.25 }&{1.00 $\pm$ 0.00} & {0.92 $\pm$ 0.05 } & {0.14 $\pm$ 0.22 }\\
\midrule
Flood and Echo \textit{random}& {$\oo(m)$}&{1.00 $\pm$ 0.00} & {1.00 $\pm$ 0.00 } & {0.99 $\pm$ 0.01 }&{1.00 $\pm$ 0.00} & {0.97 $\pm$ 0.04 } & {0.77 $\pm$ 0.30 }&{1.00 $\pm$ 0.00} & {0.82 $\pm$ 0.01 } & {0.01 $\pm$ 0.00 }\\
Flood and Echo \textit{fixed}& {$\oo(m)$}&{1.00 $\pm$ 0.00} & {1.00 $\pm$ 0.00 } & {1.00 $\pm$ 0.00 }&{1.00 $\pm$ 0.00} & {1.00 $\pm$ 0.00 } & {0.99 $\pm$ 0.02 }&{1.00 $\pm$ 0.00} & {1.00 $\pm$ 0.00 } & {1.00 $\pm$ 0.00 }\\
\bottomrule
\end{tabular}}
\label{tab:algorithmic_tasks}
\end{table*}

We test performance on the different Flood and Echo Net modes: \textit{fixed, random} and \textit{all}. All modes execute two phases, which results in $\oo(m)$ messages exchanged per chosen origin. Moreover, we choose the marked nodes in the tasks for the origin in the \textit{fixed} mode. Note that the \textit{all} mode, requires $n$ executions, one for each node, therefore, we only consider it for graphs of size at most one hundred. Nevertheless, the other modes can scale more easily and we believe them to be better suited for the study of algorithm learning. 
As a baseline comparison, we consider three models also used later on in the SALSA-CLRS evaluation. Most importantly, their architectures should be scalable to larger graph sizes and should operate on the original graph topology. We consider GIN as a representative of a maximal expressive MPNN which executes a fixed number of rounds. More precisely, five rounds are executed as the model begins to destabilize for more rounds.

\begin{corollary}
\label{cor:const_rounds}
    Let $D$ be the diameter of the Graph. Every MPNN that correctly solves the PrefixSum, Distance, or Path Finding for all graph sizes $n$ must execute at least $\oo(D)$ rounds and exchange $\oo(mD)$ messages.
\end{corollary} 

Due to the above corollary, we also consider two recurrent baselines, which adapt the number of rounds according to the graph size. Therefore, we consider RecGNN \citet{grötschla2022learning} and PGN \citep{veličković2020pointer}. We scale the number of rounds by 1.2$n$, where $n$ denotes the number of nodes in the graph.

Throughout all experiments, all models are trained on small graphs of size 10. Then, we test how well they have learned the underlying algorithmic pattern and evaluate how they generalize to graphs of size 100. From the results in Table \ref{tab:algorithmic_tasks}, we can observe that the baseline using a fixed number of layers already struggles to fit the training data and deteriorates when tested on larger instances. Similarly, the performance of PGN drops for larger graphs. The other models exhibit better generalization, especially the node accuracy remains high. 
Further, we also report the graph accuracy, which measures how many graph instances are predicted correctly, meaning that all nodes within the graph are assigned their correct label. There, we can see that the overall model performance of the baselines drops compared to the fixed variant of Flood and Echo Net.

Moreover, we can test extrapolation to even larger instances, as shown in Figure \ref{fig:prefix-plot}. Note that even though the node accuracy for many entries is quite high, the graph accuracy deteriorates as the graph sizes increase. The Flood and Echo models seem to be more robust to this phenomena, especially for the fixed origin variant. This indicates that the proposed algorithmic alignment is beneficial for size generalization. 

We continue testing The Flood and Echo Net on more challenging algorithmic tasks. Note, that our focus lies on the impact of the algorithmic alignment of our method on the study of size generalization. This is of special interest for algorithm learning, whereas we expect the applications and effects of the Flood and Echo Net to be more limited for other graph applications where the alignment with standard message-passing is more suitable. 

\subsection{SALSA - CLRS}

As demonstrated in the previous section, the Flood and Echo Net shows that it can help generalize to larger graph instances on simple algorithmic tasks. Here, we further test if the Flood and Echo Net can do so on more challenging tasks and learn more complex graph algorithms. For this purpose, we use the SALSA-CLRS benchmark \citep{minder2023salsaclrs}, which consists of a collection of six graph algorithms from the CLRS \citep{pmlr-v162-velickovic22a} collection with an emphasis on sparsity and scalability. 

\begin{wrapfigure}{r}{0.5\textwidth}
    \centering
  \includegraphics[width=\linewidth]{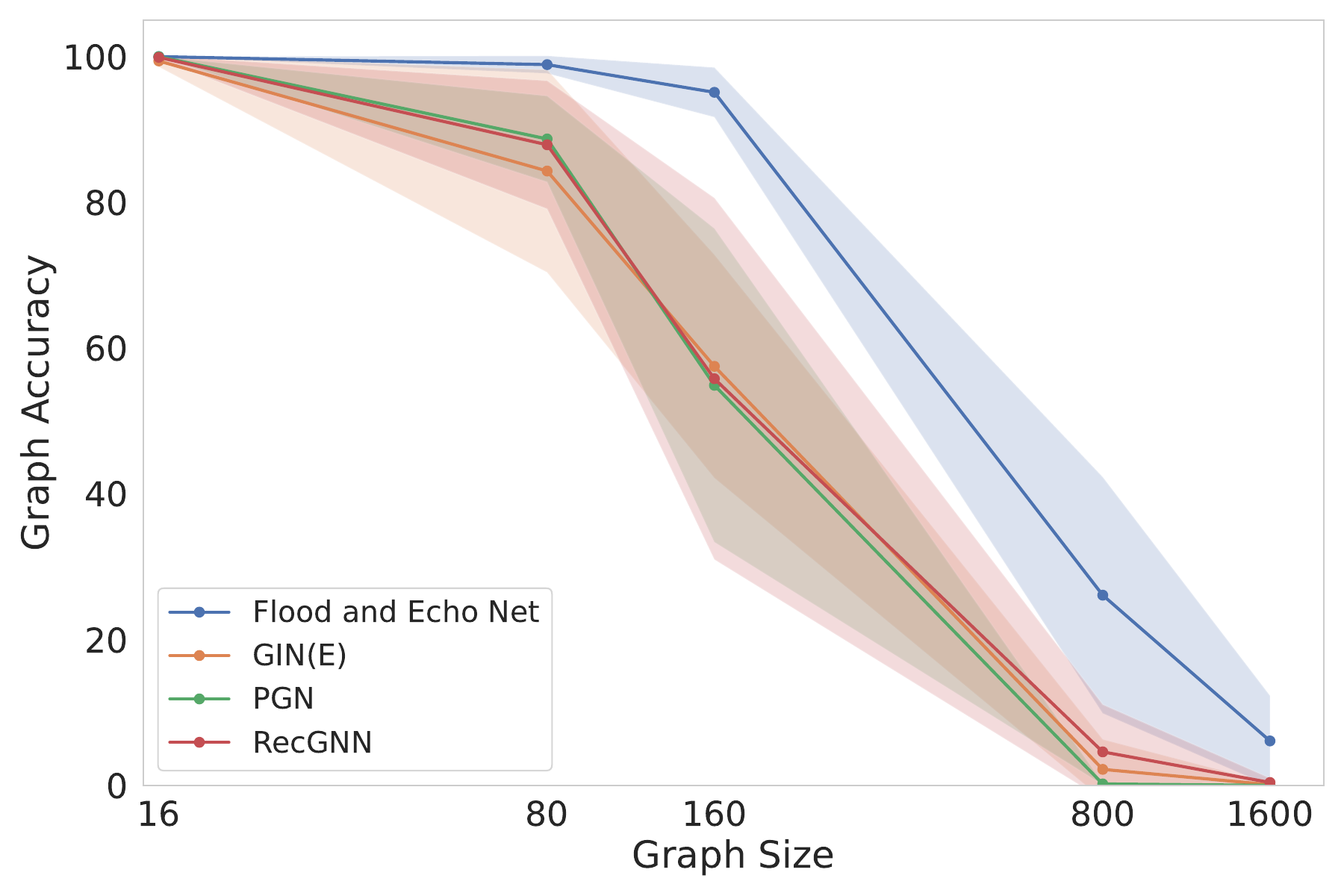}
  \caption{ Graph Accuracy on the SALSA-CLRS benchmark for the BFS task on Erd\H{o}s--R\'enyi graphs. The Flood and Echo Net can generalize almost perfectly to graphs 10 times larger while the baselines already have a significant drop in performance. Moreover, it does not rely on the number of steps given by the hints and executes a single phase.}
  \label{fig:graph_accuracy}
%\vspace{-1cm}
\end{wrapfigure}
We use the fixed variant of the Flood and Echo Net and choose the origin to match the starting node $s$ provided by the SALSA-CLRS data whenever possible, i.e. in the Dijkstra or BFS task. Otherwise, we choose the node with id 0 to be the origin. For all runs, except where explicitly stated otherwise, we do not use hints during training and run a single phase of Flood and Echo Net. Note that compared to the other baselines, the Flood and Echo Net, therefore, does not rely on being given the number of steps to be executed. 

\begin{figure}[t]
  \centering
  \includegraphics[width=0.49\linewidth]{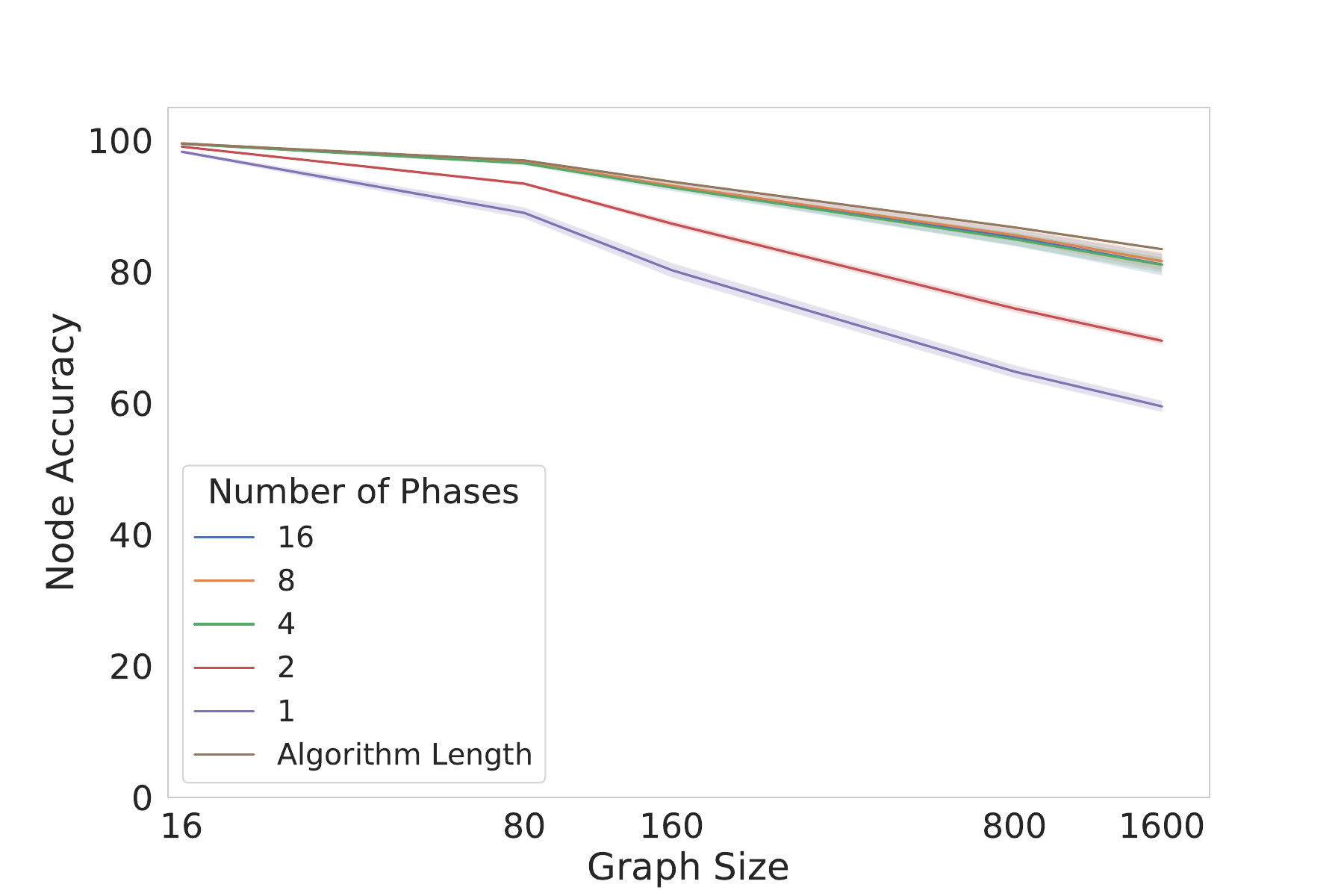}
  \includegraphics[width=0.49\linewidth]{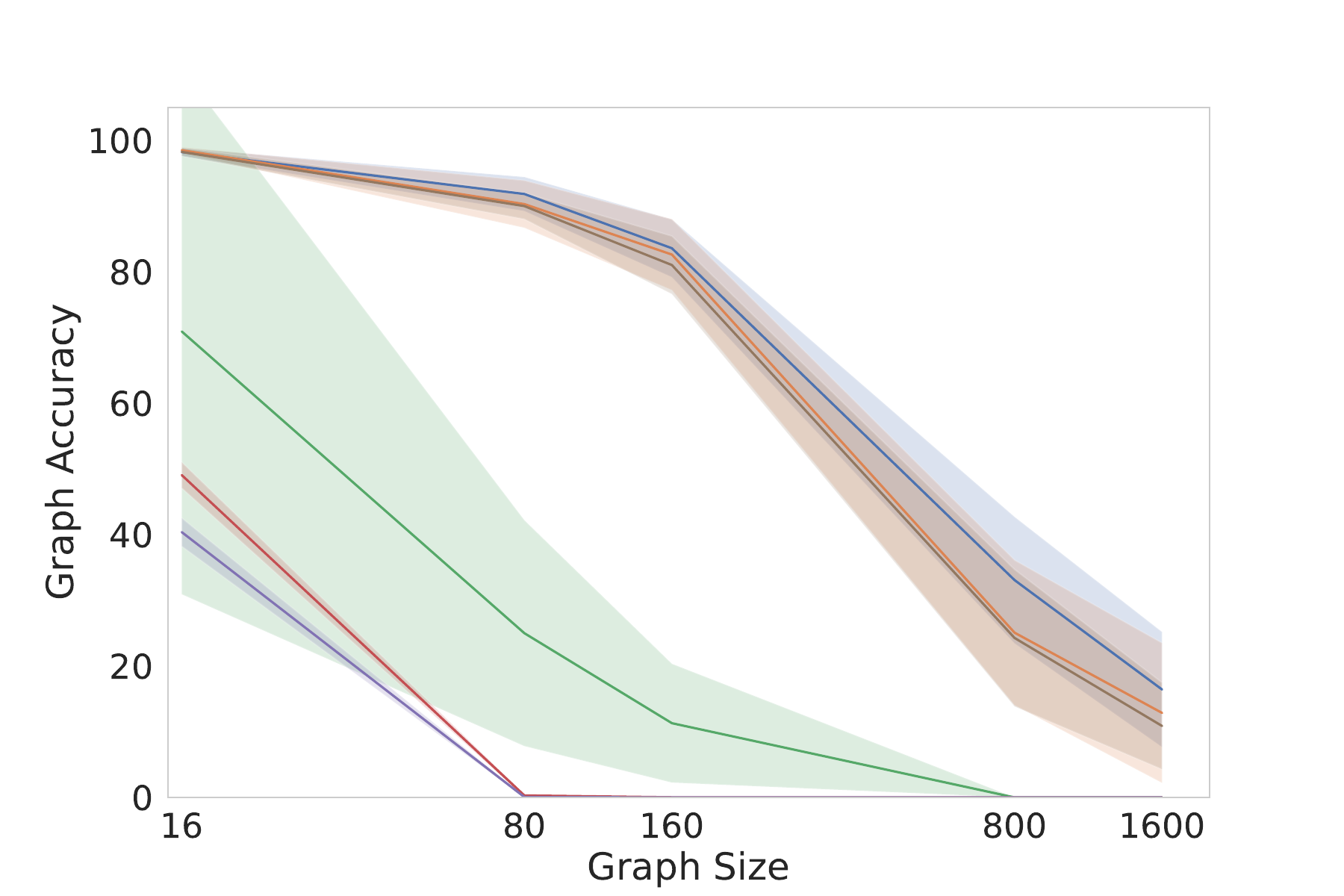}
  \caption{On the left side, we illustrate node accuracy for Dijkstra, and on the right side, graph accuracy for MIS. Adjusting the number of phases can have a positive impact on both node and graph accuracy. All models are run on Erd\H{o}s--R\'enyi graphs for a different amount of phases, Algorithm Length indicates that the number of phases is set equal to the given algorithm sequence length. }
  \label{fig:dijkstra_rounds}
\end{figure}

All models are trained on graphs of size at most 16 and then tested on larger graph sizes. In Table \ref{tab:node_accuracy}, we report the mean node accuracy and standard deviation across 5 runs. The baseline performances were taken from \citet{minder2023salsaclrs}, For further details on the technical setup, we refer to Appendix \ref{app:salsa}. The Flood and Echo Net achieves good performance across the algorithms. Most notably, the BFS and Eccentricity task can benefit from the algorithmic alignment. This is further underlined when we also look at the graph accuracies achieved on this task in Figure \ref{fig:graph_accuracy}. Here, the Flood and Echo Net achieves almost perfect scores on graphs up to size 160, while the baselines already experience a significant drop off.
For other tasks, the results are competitive with other baselines, but do not beat them with one phase of Flood and Echo.
To further investigate this, we run an additional ablation on the Dijkstra and MIS task illustrated in Figure \ref{fig:dijkstra_rounds} and find that the performance increases when the number of phases is increased. For the complete results, we refer to Appendix \ref{app:salsa}. 

Overall, the Flood and Echo Net achieves strong performance on the SALSA-CLRS benchmark. While working well on some algorithms, Flood and Echo might not be the most effective for all of them.
On some tasks, we observe an increase in performance when more phases are executed. This offers a suitable solution for learning algorithms that might not be as well aligned with the flooding and echo paradigm. However, even without relying on a given number of steps or intermediate hints, it achieves significantly better results for two of the tasks, which also translates to graph accuracy, demonstrating that it can improve generalization.

\begin{table}[t]
    \centering
    \caption{We test the Flood and Echo Net on the SALSA-CLRS benchmark across six graph-based algorithmic tasks. All models are trained on graphs of size 16 and then tested on larger graph sizes. We report the node accuracy over 5 runs on Erd\H{o}s--R\'enyi graphs of different sizes. Using a single phase of the Flood and Echo Net achieves good performance, especially on the BFS and Eccentricity task on which it exhibits strong generalization.\\}
    
    \begin{minipage}{0.48\textwidth}
        \centering
        
\resizebox{1.0\linewidth}{!}{

        \begin{tabular}{@{}l*{17}{S[table-format=-3.4]}@{}}
            \toprule
            {Task} & {Model} & {16} & {80} & {160} & {800} & {1600}\\
\midrule
 BFS & {Flood and Echo} & {{100.0 \text{\tiny $\pm$ 0.0} }} & {{100.0 \text{\tiny $\pm$ 0.0} }} & {{100.0 \text{\tiny $\pm$ 0.0} }} & {{99.7 \text{\tiny $\pm$ 0.1} }} & {{99.6 \text{\tiny $\pm$ 0.2} }} \\
 & {GIN(E)} & {100.0 \text{\tiny $\pm$ 0.1}} & {99.6 \text{\tiny $\pm$ 0.4}} & {99.3 \text{\tiny $\pm$ 0.6}} & {98.0 \text{\tiny $\pm$ 1.6}} & {98.0 \text{\tiny $\pm$ 1.5}} \\
 & {PGN} & {100.0 \text{\tiny $\pm$ 0.0}} & {99.8 \text{\tiny $\pm$ 0.1}} & {99.5 \text{\tiny $\pm$ 0.3}} & {99.0 \text{\tiny $\pm$ 0.2}} & {98.9 \text{\tiny $\pm$ 0.2}} \\
 & {RecGNN} & {100.0 \text{\tiny $\pm$ 0.0}} & {99.8 \text{\tiny $\pm$ 0.1}} & {99.5 \text{\tiny $\pm$ 0.3}} & {99.3 \text{\tiny $\pm$ 0.4}} & {99.2 \text{\tiny $\pm$ 0.4}} \\
\midrule
 DFS & {Flood and Echo} & {{69.5 \text{\tiny $\pm$ 0.7} }} & {{43.1 \text{\tiny $\pm$ 0.4} }} & {{31.5 \text{\tiny $\pm$ 0.3} }} & {{28.6 \text{\tiny $\pm$ 0.4} }} & {{26.2 \text{\tiny $\pm$ 0.4} }} \\
 & {GIN(E)} & {49.3 \text{\tiny $\pm$ 8.1}} & {30.6 \text{\tiny $\pm$ 4.0}} & {19.7 \text{\tiny $\pm$ 3.9}} & {18.1 \text{\tiny $\pm$ 3.8}} & {16.5 \text{\tiny $\pm$ 3.5}} \\
 & {PGN} & {74.2 \text{\tiny $\pm$ 14.0}} & {41.2 \text{\tiny $\pm$ 3.8}} & {29.9 \text{\tiny $\pm$ 2.6}} & {27.8 \text{\tiny $\pm$ 2.1}} & {25.8 \text{\tiny $\pm$ 2.1}} \\
 & {RecGNN} & {33.4 \text{\tiny $\pm$ 14.5}} & {28.0 \text{\tiny $\pm$ 6.5}} & {18.7 \text{\tiny $\pm$ 4.1}} & {18.2 \text{\tiny $\pm$ 4.4}} & {16.8 \text{\tiny $\pm$ 4.3}} \\
\midrule
 Dijkstra & {Flood and Echo} & {{98.3 \text{\tiny $\pm$ 0.3} }} & {{89.0 \text{\tiny $\pm$ 1.0} }} & {{80.3 \text{\tiny $\pm$ 1.4} }} & {{64.8 \text{\tiny $\pm$ 1.5} }} & {{59.6 \text{\tiny $\pm$ 1.4} }} \\
 & {GIN(E)} & {98.0 \text{\tiny $\pm$ 0.2}} & {89.8 \text{\tiny $\pm$ 1.1}} & {84.3 \text{\tiny $\pm$ 1.6}} & {75.8 \text{\tiny $\pm$ 2.2}} & {72.8 \text{\tiny $\pm$ 2.3}} \\
 & {PGN} & {99.6 \text{\tiny $\pm$ 0.1}} & {98.6 \text{\tiny $\pm$ 0.3}} & {97.2 \text{\tiny $\pm$ 0.5}} & {94.1 \text{\tiny $\pm$ 0.6}} & {92.2 \text{\tiny $\pm$ 0.7}} \\
 & {RecGNN} & {98.5 \text{\tiny $\pm$ 1.6}} & {86.8 \text{\tiny $\pm$ 15.4}} & {76.0 \text{\tiny $\pm$ 22.1}} & {63.7 \text{\tiny $\pm$ 27.7}} & {60.6 \text{\tiny $\pm$ 27.7}} \\
\bottomrule
\end{tabular}
}
    \end{minipage}
    \hfill
    \begin{minipage}{0.49\textwidth}
        \centering
        
\resizebox{1.0\linewidth}{!}{

        \begin{tabular}{@{}l*{17}{S[table-format=-3.4]}@{}}
            \toprule
            {Task} & {Model} & {16} & {80} & {160} & {800} & {1600}\\
\midrule
 Eccentricity & {Flood and Echo} & {{99.8 \text{\tiny $\pm$ 0.0} }} & {{99.2 \text{\tiny $\pm$ 0.7} }} & {{97.2 \text{\tiny $\pm$ 1.4} }} & {{98.1 \text{\tiny $\pm$ 1.5} }} & {{81.5 \text{\tiny $\pm$ 6.7} }} \\
 & {GIN(E)} & {57.3 \text{\tiny $\pm$ 21.2}} & {77.1 \text{\tiny $\pm$ 17.5}} & {72.3 \text{\tiny $\pm$ 18.0}} & {51.3 \text{\tiny $\pm$ 34.2}} & {36.7 \text{\tiny $\pm$ 17.6}} \\
 & {PGN} & {100.0 \text{\tiny $\pm$ 0.0}} & {100.0 \text{\tiny $\pm$ 0.0}} & {100.0 \text{\tiny $\pm$ 0.0}} & {100.0 \text{\tiny $\pm$ 0.0}} & {64.6 \text{\tiny $\pm$ 14.9}} \\
 & {RecGNN} & {75.8 \text{\tiny $\pm$ 26.2}} & {80.5 \text{\tiny $\pm$ 35.0}} & {75.0 \text{\tiny $\pm$ 39.1}} & {72.7 \text{\tiny $\pm$ 27.9}} & {63.0 \text{\tiny $\pm$ 24.8}} \\
\midrule
 MIS & {Flood and Echo} & {{91.5 \text{\tiny $\pm$ 0.0} }} & {{87.5 \text{\tiny $\pm$ 0.1} }} & {{87.9 \text{\tiny $\pm$ 0.1} }} & {{88.3 \text{\tiny $\pm$ 0.2} }} & {{87.4 \text{\tiny $\pm$ 0.1} }} \\
 & {GIN(E)} & {82.2 \text{\tiny $\pm$ 2.5}} & {81.6 \text{\tiny $\pm$ 1.9}} & {80.8 \text{\tiny $\pm$ 2.4}} & {83.6 \text{\tiny $\pm$ 1.5}} & {80.8 \text{\tiny $\pm$ 2.5}} \\
 & {PGN} & {99.8 \text{\tiny $\pm$ 0.1}} & {99.6 \text{\tiny $\pm$ 0.2}} & {99.5 \text{\tiny $\pm$ 0.2}} & {98.8 \text{\tiny $\pm$ 0.6}} & {98.9 \text{\tiny $\pm$ 0.5}} \\
 & {RecGNN} & {93.6 \text{\tiny $\pm$ 2.2}} & {90.0 \text{\tiny $\pm$ 2.3}} & {90.1 \text{\tiny $\pm$ 2.5}} & {87.9 \text{\tiny $\pm$ 1.9}} & {88.2 \text{\tiny $\pm$ 2.6}} \\
\midrule
 MST & {Flood and Echo} & {{87.2 \text{\tiny $\pm$ 0.3} }} & {{67.9 \text{\tiny $\pm$ 0.4} }} & {{63.5 \text{\tiny $\pm$ 0.3} }} & {{52.9 \text{\tiny $\pm$ 0.8} }} & {{48.9 \text{\tiny $\pm$ 0.8} }} \\
 & {GIN(E)} & {92.6 \text{\tiny $\pm$ 0.8}} & {79.1 \text{\tiny $\pm$ 1.3}} & {77.6 \text{\tiny $\pm$ 1.7}} & {74.5 \text{\tiny $\pm$ 2.0}} & {72.9 \text{\tiny $\pm$ 2.2}} \\
 & {PGN} & {97.3 \text{\tiny $\pm$ 0.4}} & {89.1 \text{\tiny $\pm$ 1.6}} & {84.6 \text{\tiny $\pm$ 1.7}} & {75.7 \text{\tiny $\pm$ 2.0}} & {71.9 \text{\tiny $\pm$ 2.1}} \\
 & {RecGNN} & {94.2 \text{\tiny $\pm$ 2.3}} & {70.7 \text{\tiny $\pm$ 27.8}} & {66.6 \text{\tiny $\pm$ 28.2}} & {58.9 \text{\tiny $\pm$ 29.0}} & {56.0 \text{\tiny $\pm$ 28.5}} \\
\bottomrule
\end{tabular}
}
    \end{minipage}
    \label{tab:node_accuracy}
\end{table}

\section{Conclusion}
In this work, we challenge the standard message-passing paradigm commonly used by many Graph Neural Networks and introduce the Flood and Echo Net. It aligns its execution to a design pattern from distributed algorithms where messages are flooded and echoed throughout the entire graph in a wave-like activation. Through these structured activations of the nodes, the Flood and Echo Net becomes more expressive than regular MPNNs. Moreover, our method is more efficient with regard to message complexity, as it can facilitate messages throughout the graph more easily. Finally, the execution of the Flood and Echo Net naturally generalizes to graphs of larger sizes, which we find to be helpful in improving generalization in algorithm learning. We empirically validate our findings on simple algorithmic tasks as well as more challenging graph algorithms from the SALSA-CLRS benchmark. There, we find that the algorithmic alignment of the Flood and Echo Net significantly increases performance on at least two of the algorithms in both the node and graph accuracy. These results underline that the Flood and Echo Net can improve generalization. 

%%%%%%%%%%%%%%%%%%%%%%%%%%%%%%%%%%%%%%%%%%%%%%%%%%%%%%%%%%%%

\bibliographystyle{abbrvnat} % or any other bibliography style you prefer
\bibliography{neurips_2024} % replace "references" with the name of your .bib file without the extension

\appendix

\section{Flood and Echo Net Definition}
\label{app:flood_and_echo}
Let $r$ be the origin of the computation phase and let $d(v)$ denote the shortest path distance from $v$ to $r$. 
Then, the update rule for of the Flood and Echo Net looks is defined as follows, assume $T$ phases are executed. At the beginning of each phase $t$, the flooding is performed, where the nodes are sequentially activated one after another depending on their distance towards the root. Each convolution is either from nodes at distance $d$ to $d+1$ (flood), from $d+1$ to $d$ (echo) or between nodes at the same distance (floodcross, echocross). The term $x[d]$ denotes that only nodes at distance $d$ update their state. For each distance $d$ from 1 to the max distance in the graph the following update is performed: 
\begin{align*}
    f_v^t &= \text{AGGREGATE}_{Flood}(\{\!\{x_u^t \mid d(u) = d - 1, u\in N(v)\}\!\})\\
    x_v^{t+1}[d] &= \text{UPDATE}_{Flood}(x_v^t, f_v^t)\\
    fc_v^t &= \text{AGGREGATE}_{FloodCross}(\{\!\{x_u^{t+1} \mid d(u) = d, u \in N(v)\}\!\})\\
    x_v^{t+1}[d] &= \text{UPDATE}_{FloodCross}(x_v^{t+1}, fc_v^t)\\
\end{align*}

\begin{figure}
\centering
%\framebox[4.0in]{$\;$}
\includegraphics[scale=0.70]{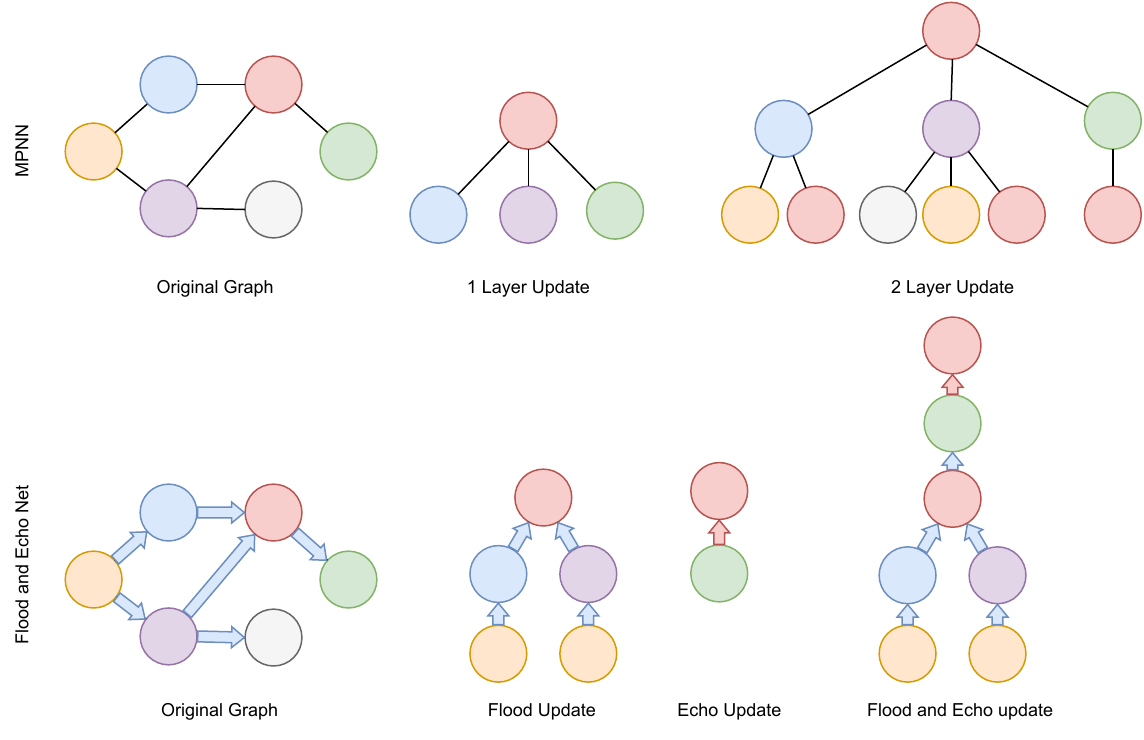}
\caption{Visualization of the computation executed on the same graph for a regular MPNN and a Flood and Echo Net from the perspective of the red node. The top row shows the computation for regular MPNN both for 1 and 2 layers of message-passing. Note that executing $l$ layers takes into account the $l$-Hop neighborhood. On the bottom row, the computation from the perspective of the red node in a Flood and Echo net is shown. Note that the origin of the Flood and Echo Net is the orange node. The two middle figures illustrate the updates in the flood and the echo part respectively. Furthermore, the figure on the right shows the combined computation for an entire phase.
}
\label{fig:computation_tree}
\end{figure}

And similarly for each distance $d$ from max distance -1 to 0 the Echo phase
\begin{align*}
    ec_v^t &= \text{AGGREGATE}_{EchoCross}(\{\!\{x_u^{t} \mid d(u) = d+1, u \in N(v)\}\!\})\\
    x_v^{t+1}[d] &= \text{UPDATE}_{EchoCross}(x_v^{t}, ec_v^t)\\
    e_v^t &= \text{AGGREGATE}_{Echo}(\{\!\{x_u^{t+1} \mid d(u) = d, u \in N(v)\}\!\})\\
    x_v^{t+1}[d] &= \text{UPDATE}_{Echo}(x_v^{t+1}, e_v^t)\\
\end{align*}
The phase is completed after another update for all nodes.
\begin{align*}
    x_v^{t+1} &= \text{UPDATE}(x_v^{t+1})
\end{align*}
Note that the node activations are done in a sparse way, therefore, for all updates that take an empty neighborhood set as the second argument no update is performed and the state is maintained. Furthermore, in practise we did not find a significant difference in performing the last update step, which is why in the implementation we do not include it. 
In Figure \ref{fig:computation_tree} we outline the differences between the computation of an MPNN and a Flood and Echo Net.

%\section{Flood and Echo Net}

\section{Extended Related Work}
A variety of GNNs that do not follow the 1 hop neighborhood aggregation scheme have been unified under the view of so-called Subgraph GNNs. The work of ~\citet{zhang2023complete} analyses these models in terms of their expressiveness and gives the following general definition:

\begin{definition}
\label{def:layer}
A general subgraph GNN layer has the form
\begin{equation*}
    h_G^{(l+1)}(u, v) = \sigma^{(l+1)}(\op_1(u, v, G, h_G^{(l)}),\cdots ,\op_r(u, v, G, h_G^{(l)})),
\end{equation*}
where $\sigma^{(l+1)}$ is an arbitrary (parameterized) continuous function, and each atomic operation $\op_i(u,v,G,h)$ can take any of the following expressions:
\begin{itemize}
\raggedright
\setlength{\itemsep}{0pt}
    \item Single-point: $h(u,v)$, $h(v,u)$, $h(u,u)$, or $h(v,v)$;
    \item Global: $\sum_{w\in\gV_G}h(u,w)$ or $\sum_{w\in\gV_G}h(w,v)$;
    \item Local: $\sum_{w\in\gN_{G^u}(v)}h(u,w)$ or $\sum_{w\in\gN_{G^v}(u)}h(w,v)$.
\end{itemize}
We assume that $h(u,v)$ is always present in some $\op_i$.
\end{definition}
This allows us to capture a more general class of Graph Neural Networks, i.e., the work of ~\citet{zhang2023rethinking}, which can incorporate distance information into the aggregation mechanism this way. 
Note that the proposed mechanism of the Flood and Echo Net differs from that of this particular notion of subgraph GNNs. At each update step, only a subset of nodes is active. This allows nodes to take into account nodes that are activated earlier, which is not directly comparable to subgraph GNNs where the node updates still happen simultaneously for the nodes in question.

Another important issue that GNNs often struggle with is the so-called phenomenon of oversquashing \citep{alon2021bottleneck}. In simple terms, if too much information has to be propagated through the graph using a few edges, a bottleneck occurs, squashing the relevant information together, leading to information loss and subsequent problems for learning. Recent work of \citep{digiovanni2023oversquashing} theoretically analyses the reasons leading to the oversquashing phenomena and identifies the width and depth of the network but also the graph topology as key contributors. Note that the proposed Flood and Echo Net is not designed to tackle the problem of oversquashing. Rather, it tries to facilitate information throughout the graph, assuming that there is no inherent (topological) bottleneck. It only affects the aforementioned depth aspect of the network. However, as outlined by \citep{digiovanni2023oversquashing}, the depth is likely to have a marginal effect compared to the graph topology.

The works of \citet{martinkus2023agentbased}, namely AgentNet, and \citet{faber2023asynchronous}, who proposes AMP (Asynchronous Message Passing), also draw inspiration from the field of distributed computing. Although they share some aspects in their mechanisms, their respective settings differ quite a bit. In AgentNet, there exist agents which traverse the graph which gives them the possibility to solve problems on the graph in sublinear time. In contrast, our approach tries to enable communication throughout the whole graph, especially in the context of different graph sizes. On the other hand, AMP activates nodes one at a time, benefiting from a similar computational sparsity as our method. However, note that the Flood and Echo Net's execution is more structured. On one side, this leads to less flexible activation patterns, however, on the other hand, it translates naturally across graph sizes. Whereas AMP has to additionally learn a termination criteria which must generalize.

\section{1-WL Expressive Experiments}
\label{app:expressive_results}

We empirically validate our findings for the Flood and Echo Net on multiple expressive datasets that go beyond 1-WL. The tasks span both graph and node predictions, which include graphs that have multiple disconnected components. We test two modes on these datasets. One variant performs an execution from a single node using the random variant, while the other performs the all mode. Both modes compute node embeddings and can be used for the node prediction tasks without modification. Whereas for graph prediction tasks, the sum of all node class predictions is used for the final graph prediction. 
Note that the second variant is fairer for comparison against MPNNs, since for some datasets like Limits-1, Limits-2, and 4-Cycles, the graph is not connected. Therefore, the single start mode struggles, as it cannot access all components. 

In Table \ref{table:expressive} we can see that the Flood and Echo all starts manages to almost perfectly solve all tasks. The single start performs worse in the Limits-1 and Limits-2 due to the lack of access to all components. The GIN model, as predicted by theory, performs no better than random guessing. The higher scores in the Triangles and LCC datasets are due to the fact that these datasets are imbalanced. For an in-depth explanation of the individual datasets, we refer to Appendix \ref{appendix:expressive_datasets}.
Comparing the message complexities of the different methods, a GIN with $L$ layers exchanges $\oo(Lm)$ messages while the Flood and Echo model either exchanges $\oo(m)$ or $\oo(nm)$ messages based on whether it executes the single or all starts mode. 

\begin{table*}[t]
\centering
\caption{As the theory predicts, the GIN model cannot go beyond trivial performance.  Whereas both the \textit{single} and \textit{all} execution mode go beyond the limits of $1$-WL. Note, that the datasets are imbalanced and can contain multiple components, which can explain the performance of GIN and the account for the drop of the single mode compared to the all execution.}
\resizebox{0.9\textwidth}{!}{
\begin{tabular}{@{}l*{7}{S[table-format=-3.4]}@{}}
\toprule
{Model} & \multicolumn{2}{c}{\textsc{GIN}} & \multicolumn{2}{c}{\textsc{Flood and Echo } \textit{single}} & \multicolumn{2}{c}{\textsc{Flood and Echo } \textit{all}}\\
{}  & {Train} & {Test} & {Train} & {Test} & {Train} & {Test} \\
\midrule  
Triangles &{0.80 $\pm$ 0.00} & {0.78 $\pm$ 0.00 }&{0.92 $\pm$ 0.00} & {0.92 $\pm$ 0.00 }&{1.00 $\pm$ 0.00} & {1.00 $\pm$ 0.00 }\\
LCC &{0.79 $\pm$ 0.00} & {0.79 $\pm$ 0.00 }&{0.92 $\pm$ 0.00} & {0.91 $\pm$ 0.00 }&{1.00 $\pm$ 0.00} & {1.00 $\pm$ 0.00 }\\
4-Cycles &{0.49 $\pm$ 0.02} & {0.50 $\pm$ 0.00 }&{0.95 $\pm$ 0.01} & {0.95 $\pm$ 0.02 }&{1.00 $\pm$ 0.00} & {0.96 $\pm$ 0.02 }\\
Limits-1 &{0.50 $\pm$ 0.00} & {0.50 $\pm$ 0.00 }&{0.70 $\pm$ 0.06} & {0.80 $\pm$ 0.27 }&{1.00 $\pm$ 0.00} & {1.00 $\pm$ 0.00 }\\
Limits-2 &{0.50 $\pm$ 0.00} & {0.50 $\pm$ 0.00 }&{0.79 $\pm$ 0.05} & {0.90 $\pm$ 0.22 }&{1.00 $\pm$ 0.00} & {1.00 $\pm$ 0.00 }\\
Skip-Circles &{0.10 $\pm$ 0.00} & {0.10 $\pm$ 0.00 }&{1.00 $\pm$ 0.00} & {1.00 $\pm$ 0.00 }&{1.00 $\pm$ 0.00} & {1.00 $\pm$ 0.00 }\\
\midrule
Messages & \multicolumn{2}{c}{$\oo(Lm)$} & \multicolumn{2}{c}{$\oo(m)$} & \multicolumn{2}{c}{$\oo(nm)$}\\
\bottomrule
\end{tabular}}
\label{table:expressive}
\end{table*}

%===================
% Information Propagation
%===================
\section{Information Propagation}
\label{app:inf_prop}

\begin{figure}
\centering
\includegraphics[width=0.9\linewidth]{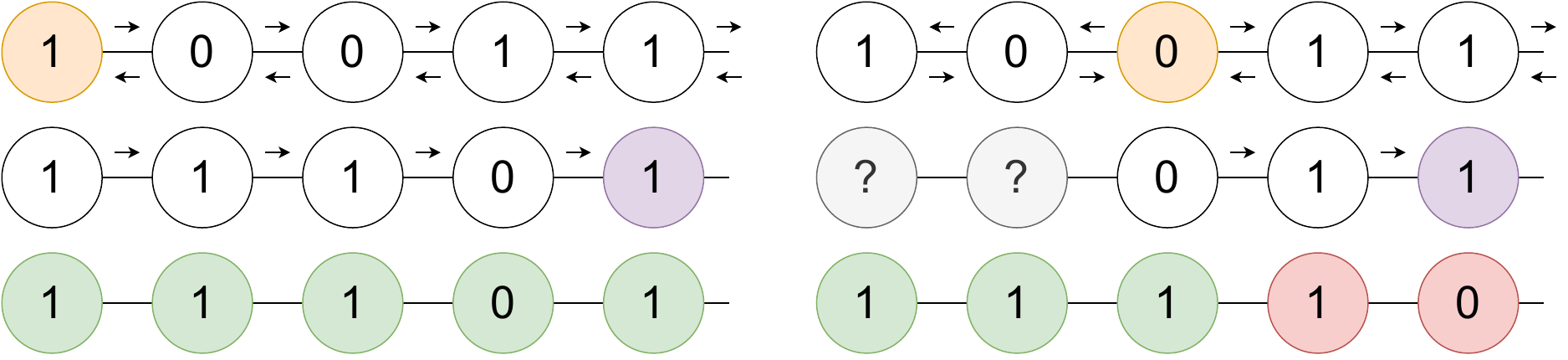}
\caption{Visualization of the information exchange in the PrefixSum task when choosing different origin nodes for Flood and Echo Net. We can derive theoretical upper bounds for the performance of Flood and Echo Net depending on the number of random origin nodes for a single phase. We show that the empirical performance closely follows the theoretical analysis. This confirms the ability of the Flood and Echo Net to distribute the available information throughout the whole graph.}
\label{fig:prefix_information}
\end{figure}
In this section, we analyze the ability of the Flood and Echo Net to distribute the available information throughout the whole graph. We use a synthetic algorithmic dataset, the \textsc{PrefixSum} task. For this task, we can provably determine what pieces of information must be gathered for each node to make correct predictions. If we choose an appropriate origin point, we could easily send the information and solve the task. However, more interestingly, what happens if we choose a random origin node instead? Can the Flood and Echo model still distribute the relevant information, even if it does not suffice to fully solve the task? We derive theoretical upper bounds for the best-performing instance given the information that theoretically could be available during the execution depending on the number of origin nodes. Interestingly, even if the full information is not available, the Flood and Echo Net achieves performance that closely follows the theoretical upper bound. This showcases the ability of our proposed method to distribute all available information throughout the whole graph.

\begin{table*}[t]
\centering
\caption{Information propagation of the Flood and Echo Net for graphs of size $n$ on the PrefixSum task. As the number of random origin points $s$ increases, the model can distribute the additional information, as seen by the increase in accuracy. Moreover, it can do so very effectively as the performance closely follows the theoretical upper bound. }
\resizebox{\textwidth}{!}{
\begin{tabular}{@{}l*{9}{S[table-format=-3.4]}@{}}
\toprule
{Model} & \multicolumn{4}{c}{$n=10$} & \multicolumn{4}{c}{$n=100$}\\
{} & {$s=1$} & {$s=2$} & {$s=3$} & {$s=5$} & {$s=1$} & {$s=2$} & {$s=3$} & {$s=5$} \\
\midrule  
{\textsc{Theoretical Upper Bound}} & 82.00 & 89.80 & 93.52 & 96.91 & 75.75 & 84.07 & 88.23 & 92.39 \\
%\midrule
{\textsc{Flood and Echo}} & { 81.69 $\pm$ 0.51 } & { 88.10 $\pm$ 2.34 } & { 89.99 $\pm$ 0.28 } & { 93.90 $\pm$ 0.23 } & { 75.39 $\pm$ 0.29 } & { 83.43 $\pm$ 0.44 } & { 87.79 $\pm$ 0.34 } & { 91.86 $\pm$ 0.28 } \\
\bottomrule
\end{tabular}
}
\label{table:prefix_information}
\end{table*}

\paragraph{PrefixSum Task}
\label{intro_prefixsum}
For this analysis, we use the PrefixSum dataset, which follows the task introduced by \citet{NEURIPS2021_3501672e} and was later adapted for the graph setting \citep{grötschla2022learning}. It consists of a path graph, where one end is marked to distinguish left form right. Each node $v$ independently and uniformly at random gets assigned one bit $x_v$, which is either $1$ or $0$, chosen with probability $\frac{1}{2}$ each. The task is to compute the prefix sum from left to right modulo 2. Therefore, the output $y_v$ of each node $v$ is the sum of the bits of all nodes to the left $y_v \equiv_2 (\sum_{i\leq v}x_i)$. 
Note, that in order to correctly predict a node output, it has to take all bits left of it into consideration. 
\begin{lemma}
\label{lemma:prefix_information}
    In the PrefixSum task, for every node $v$, the computation of the output $o_v$ must be dependent on all bits of the nodes to its left. If not all bits are considered for the computation, the probability of a correct prediction is bounded by $\Pr[o_v = y_v] \leq \frac{1}{2}$.  
\end{lemma}

Note that from this lemma, it immediately follows that to solve the task correctly, information needs to be exchanged throughout the whole graph. Nodes towards the end of the path must consider almost all nodes of the graph for their computation.

\begin{corollary}
\label{cor:prefix_sum_od}
    The PrefixSum task requires information of nodes that are $\oo$(D) hops apart and therefore must exchange information throughout the entire graph. 
\end{corollary}

From Lemma \ref{lemma:prefix_information}, we know that nodes can only correctly predict their output if the information of all nodes left to them is taken into account. Whenever the initial origin of the Flood and Echo Net is chosen at one of the ends, this information should be available in either the flooding or echo part. However, what happens if we choose one of the nodes in the graph at random to be the origin? Then, there will always be a right side whose predictions are dependent on the computation of the left, which has not yet been exchanged. An example is depicted in Figure \ref{fig:prefix_information}.
The top row indicates the origin node (orange) and illustrates the message exchange in the flooding (top arrows) and echo phase (bottom arrows). The middle row indicates what parts of the graph the purple-marked node can know about after a single phase. Note that on the right-hand side, it cannot infer the initial features of the two leftmost nodes. Because of the missing information, the configuration on the right can only correctly predict the nodes up to the initial origin node.

We can formally derive a theoretical upper bound for the expected number of correctly predicted nodes depending on $n$, the number of nodes, and $s$, the number of origins. For the entire derivation and formula, we refer to the Appendix.

In Table \ref{table:prefix_information}, we can compare the empirical performance of the Flood and Echo Net with the theoretical upper bound. Moreover, the measured performance closely follows the theoretical upper bound. The experiment clearly shows that the accuracy of the model strictly increases when more starting nodes are chosen. This indicates, that the model can make use of the additional provided information. Therefore, it can effectively incorporate the information and propagate it in a sensible way throughout the graph.

\section{Proofs and Derivations}
\label{app:proofs} 

Derivation of $\E[X]$:\\
Let us assume $s$ starting nodes are chosen uniformly at random and $s_j$ denote the index of the $j$-th starting nodes. If the beginning is chosen, then all nodes could be classified correctly. Otherwise, nodes can only be correctly inferred up to $t = \max_j s_j$, the starting node farthest to the right. Moreover, the rest of the $n-t$ nodes can only be guessed correctly with probability $\frac{1}{2}$ as the cumulative sum to the left is missing.  We can derive the closed-form solution for $X$, the expected number of correctly predicted nodes for a perfect solution.

\begin{align*}
    \E[X] &= \Pr[\min_j s_j = 1]n + (1-\Pr[\min_j s_j = 1])\sum_{i=2}^{n}\frac{n+\max_js_j}{2}\Pr[\max_j s_j = i] \\
        &= \left(1 - \left(\frac{n-1}{n}\right)^s\right)n + \left(\frac{n-1}{n}\right)^s\sum_{i=2}^{n}\frac{n+i}{2}(\Pr[\max_j s_j < i+1]-\Pr[\max_j s_j < i]) \\
        &= \left(1 - \left(\frac{n-1}{n}\right)^s\right)n + \left(\frac{n-1}{n}\right)^s \sum_{i=2}^{n}\frac{n+i}{2}\left(\left(\frac{i-1}{n-1}\right)^s-\left(\frac{i-2}{n-1}\right)^s\right) \\
\end{align*}

\renewcommand*{\proofname}{Proof of Theorem \ref{thm:flood_echo_expressiveness}}
\begin{proof}
    It has been shown by the work of \citet{xu2018powerful} that the \textit{Graph Isomorphism Network} (GIN) achieves maximum expressiveness amongst MPNN. In the following, we will show that a Flood and Echo Net can simulate the execution of a GIN on connected graphs, therefore matching it in its expressive power. 
    Let $G_I$ be a GIN using a node state vector $h^k_v$ of dimension $d_i$. 
    \begin{align*}
    h_v^{(k)} = \text{MLP}^{(k)}((1+\epsilon)h_v^{(k-1)} + \sum_{u \in \mathcal{N}(v)}h_u^{(k-1)})        
    \end{align*}

    Let $G_{F}$ be a Flood and Echo Net using node state vector $q^{(k)}_v$ of dimension $d_f = 2\cdot d_i$. We partition the vector $q^{(k)}_v = o^{(k)}_v \mid\mid n^{(k)}_v$ into two vectors of dimension $d_i$. 
    Initially, we assume that the encoder gives us $o^{(0)}_v = h_v^{(0)}$ and $n_v = 0^{d_i}$ the zero vector. 
    We now define the updates of flood, floodcross, echo, and echocross in a special way, that after the flood and echo part $o^{(k)}_v$ is equal to $h^{(k)}_v$ and $n^{(k)}_v$ is equal to $\sum_{u \in \mathcal{N}(v)}h_u^{(k-1)}$.
    If this is ensured, the final update in a flood and echo phase can update $q_v^{(k)} = \text{MLP}^{(k)}((1+\epsilon)o_v^{(k-1)} + n^{(k-1)}_v) \mid\mid 0^{d_i}$, which exactly mimics the GIN update.
    It is easy to verify that if we set the echo and flood updates to add the full sum of the $o_v^{(k)}$ part of the incoming messages (and similarly half of the sum of the incoming messages during the cross updates) to $n_v^{(k-1)}$ the desired property is fulfilled. Moreover, there are at most four messages exchanged over each edge of the graph. Specifically, four is for cross edges and two is for all other edges. Therefore, a total of $\oo(m)$ messages are exchanged, which is asymptotically the same number of messages GIN exchanges in a single update step. This enables a single phase of the Flood and Echo Net to mimic the execution of a single GIN round. Repeating this process the whole GIN computation can be simulated by the Flood and Echo Net. 

    Therefore, given a GIN network $G_I$ of width $d_i$, we can construct a Flood and Echo Net $G_F$ of width $\oo(d)$ that can simulate one round of $G_I$ in a single flood and echo phase using $\oo(m)$ messages.
    % The idea is to duplicate the d state of the twice and then do two phases of the flood echo, which requires more messages but only a constant factor, so one phase just sums up all the states and the second one can update the states

\end{proof}

\renewcommand*{\proofname}{Proof of Theorem \ref{thm:flood_echo_wl}}
\begin{proof}

    To show that the Flood and Echo Net goes beyond 1-WL, it suffices to find two different graphs that are equivalent under the 1-WL test but can be distinguished by a Flood and Echo Net. 
    Observe that a Flood and Echo Net can calculate its distance, in number of hops, to the root for each node. See the graphs illustrated in Figure \ref{fig:expressive} for a comparison. On the left is a cycle with 11 nodes, which have additional connections to the nodes that are at distance two away. Similarly, the graph on the right has additional connections at a distance of three. Both graphs are four regular and can, therefore, not be distinguished using the 1-WL test. However, no matter where the starting node for Flood and Echo is placed, it can distinguish that there are nodes which have distance four to the starting root in one graph, which is not the case in the other graph. Therefore, Flood and Echo Net can distinguish the two graphs and is more expressive than the 1-WL test. Moreover, due to the Theorem \ref{thm:flood_echo_expressiveness} it matches the expressiveness of the 1-WL test on connected graphs by a reduction to the graph isomorphism network.
\end{proof}

\renewcommand*{\proofname}{Proof of Lemma \ref{lemma:flood_echo_messages}}
\begin{proof}
    Consider either one of the PrefixSum, Distance, or Path Finding tasks presented in Appendix \ref{appendix:algorithmic_datasets}. All of them require information that is $\oo(D)$ apart and must be exchanged. It follows that all MPNNs must execute at least $\oo(D)$ rounds of message-passing to facilitate this information. Moreover, in these graphs, the graph diameter can be $\oo(n)$. As in each round, there are $\oo(m)$ messages exchanged, MPNNs must use at least $\oo(nm)$ messages to solve these tasks. Furthermore, from Lemma \ref{lemma:flood_echo_solve}, it follows that Flood and Echo Net can solve the task in a single phase using $\oo(m)$ messages. 
\end{proof}

\renewcommand*{\proofname}{Proof of Lemma \ref{lemma:prefix_information}}
\begin{proof}
    For the sake of contradiction, assume that not all bits of the nodes to the left have to be considered for the computation. Therefore, at least one bit at a node $u$ exists, which is not considered for the computation of $o_v$. We know that all bits $x$ are drawn uniformly at random and are independent of each other. Furthermore, we can rewrite the groundtruth $y_v  \equiv_2  \sum_{i\leq v}x_i \equiv_2 x_u + \sum_{i\leq v, i \neq u}x_i \equiv_2 x_u + s$ as the sum of $x_u$ and the rest of the nodes. From there, it follows that the ground truth is dependent on $x_u$, even if all other bits are known $\Pr[y_v = 0 \mid s] = \Pr[s = x_u] = \frac{1}{2}$. On the other hand, we know that $o_v$ must be completely determined by the information of the nodes that make up $s$ and cannot change depending on $x_u$. Therefore, $\Pr[o_v = y_v \mid o_v \text{ does not consider } x_u] \leq \frac{1}{2}$. 
\end{proof}

\renewcommand*{\proofname}{Proof of Corollary \ref{cor:prefix_sum_od}}
\begin{proof}
    According to Lemma \ref{lemma:prefix_information}, for each node $v$ to derive the correct prediction, all $x_u$ for nodes $u$ that are left of $v$ have to be considered. Therefore, look at the node $r$ on the very right end of the path graph. It has to take the bits of all other nodes into consideration. However, the leftmost bit at node $l$ is $n-1$ hops away, which is also the diameter of the graph. Therefore, in order to solve the PrefixSum task, information has to be exchanged throughout the entire graph by propagating it for at least $\oo(D)$ hops.
\end{proof}

\renewcommand*{\proofname}{Proof of Corollary \ref{cor:distance_od}}
\begin{proof}
For the task PrefixSum, the statement follows from \ref{cor:prefix_sum_od}. For the other tasks we outline the proof as follows:
Assume for the sake of contradiction that this is not the case and only information has to be exchanged, which is $d' = o(D)$ hops away to solve the task. Therefore, as both tasks are node prediction tasks, the output of each node is defined by its $d'$-hop neighborhood. For both tasks, we construct a star-like graph $G$, which consists of a center node $c$ and $k$ paths of length $\frac{n}{k}$, which are connected to $c$ for a constant $k$. For the Path Finding task, let the center $c$ be one marked node, and the end of path $j$ be the other marked node. Consider the nodes $x_i$, $i = 1, 2, ..., k$ which lie on the $i$-th path at distance $\frac{n}{2k}$ from $c$. Note that all $x_i$ are $\frac{n}{2k}$ away from both their end of the path and $c$ the root. Moreover, the diameter of the graph is $\frac{2n}{k}$. This means that neither the end of the $i$-th path nor the center $c$ will ever be part of the $d'$hop neighborhood. Therefore, if we can only consider the $d'$-hop neighborhood for each $x_i$, they are all the same and as a consequence will predict the same solution. However, $x_j$ lies on the path between the marked nodes while the other $x_i$'s do not. So they should have different solutions, a contradiction.
A similar argument holds for the Distance task. Again let $c$ be the marked node in the graph and $x_i$ for $i=1, 2, ..., k$ be the nodes which lie on the $i$-th path at distance $\frac{n}{2k}$ for even $i$ and $\frac{n}{2k}+1$ for odd $i$. Again, note that the $d'$-hop neighborhood of all $x_i$ is identical and therefore must compute the same solution. However, the solution of even $x_i$ should be different from the odd $x_i$, a contradiction.      
\end{proof}

\renewcommand*{\proofname}{Proof of Corollary \ref{cor:const_rounds}}
\begin{proof}
From Corollary \ref{cor:distance_od} and \ref{cor:prefix_sum_od} it directly follows that information must be exchanged for at least $\oo(D)$ hops to infer a correct solution. As MPNNs only exchange information one hop and exchange $\oo(m)$ messages per round, the claim follows immediately.  
\end{proof}

\begin{lemma}
    \label{lemma:flood_echo_solve}
    Flood and Echo Net can facilitate the required information for the PrefixSum, Distance and Path Finding task in a single phase, which can be executed using $\oo(m)$ messages. 
\end{lemma}

\renewcommand*{\proofname}{Proof of Lemma \ref{lemma:flood_echo_solve}}
\begin{proof}
    We will prove that in all three mentioned tasks, there exists a configuration for a Flood and Echo phase, which can propagate the necessary information throughout the graph in a single phase. Let the starting point $s$ correspond to the marked node in the graph, or in the case of the Path Finding, any of the two suffices. First, we consider the PrefixSum task. Note that in the flooding phase, information is propagated from the start, which is the left end, towards the right. Therefore, in principle, each bit can be propagated to the right, which suffices to solve the task according to \ref{lemma:prefix_information}. For the Distance task, it is necessary that the length of the shortest path between the root and each node can be inferred. Note that this is exactly the path which is taken by the flooding messages, therefore, this should be sufficient to solve the task. Similarly, for the Path Finding task, one phase is sufficient. Note that starting from the leaves of the graph during the echo phase, nodes can decide that they are not part of the path between the two marked nodes (as only marked leaves can be part of the path). However, when such a message is received at one of the marked nodes, they can ignore it and tell their predecessor that they are on the path. This is correct, as one of the marked ends is at the start of our computation, and this echo message travels from the other marked end on the to-be-marked path back toward the root. This shows that for each of the above-mentioned tasks, there exists a Flood and Echo Net configuration that solves the task in a single phase, which exchanges $\oo(m)$ messages. 
\end{proof}

\section{Model Architecture and Training}

The following describes the setup of our experiments for PrefixSum, Path-Finding and Distance. We use a GRUMLP convolution for all Flood and Echo models and the RecGNN, which is defined in equation \ref{eqn:GRUMLP}. It concatenates both endpoints of an edge for its message and passes it into a GRU cell \citep{cho2014properties}. All models use a hidden node state of $32$. We use a multilayer perceptron with a hidden dimension $4$ times the input dimension and map back to the hidden node state. Further, we use LayerNorm introduced by \citep{ba2016layer}. We also adapt the PGN for the experiments following the implementation by \citet{minder2023salsaclrs}. We concatenate the current, last and original input in each step and also adapt the number of rounds to be linear in the graphs size by executing $1.2n$ rounds. For the expressiveness tasks, we perform one phase of Flood and Echo to compute our node embeddings, while for the algorithmic tasks, we perform two phases of Flood and Echo. We run for a maximum of 200 epochs, but do an early stop whenever the validation loss does not increase for 25 epochs. We use the Adam optimizer with an initial learning rate of 

\begin{align}
    x_v^{t+1} = \text{GRUCell}\left(x_v^t,\sum_{u\in N(v)}{\phi(x_v^t||x_u^t)}\right)
\label{eqn:GRUMLP}
\end{align}

In all our experiments, we train our model using the ADAM optimizer \citet{KingmaB14} with a learning rate of $4\cdot10^{-4}$ and batch size of $32$ for $200$ epochs. We also use a learning rate scheduler where we decay the learning rate with patience of $3$ epochs and perform early stopping if the validation loss does not decrease for more than 25 epochs. 
All reported values are reported over the mean of $5$ runs. 

The model is implemented in pytorch lightning using the pyg library, and the code will be made public upon publication. 

\section{Runtime}
\label{app:runtime}
\subsection{Runtime Complexity}

We denote $n$ the number of nodes, $m$ the number of edges and $D$ the diameter of the graph. Furthermore, let $T$ be the number of phases for a Flood and Echo Net and $L$ be the number of layers for an MPNN.

A single round of regular message-passing exchanges $\oo(m)$ messages. Therefore, executing $L$ such rounds results in $\oo(L)$ steps and $\oo(Lm)$ messages. Note that in order for communication between any two nodes $L$ has to be in the order of $\oo(D)$.

A single phase of a Flood and Echo Net, consisting of one starting node, exchanges $\oo(m)$ messages and does so in $\oo(D)$ steps. Therefore, executing $T$ phases of a Flood and Echo Net results in $\oo(Tm)$ messages exchanged in $\oo(TD)$ steps. Note, that it is sufficient for $T$ to be constant $\oo(1)$ in order to communicate throughout the whole graph and does not necessarily have to be scaled according to the size of the graph.
 
The variations \textit{fixed} and \textit{random} perform their executions only for a specific single node. Contrary, the \textit{all} variation performs such an execution for each of its nodes individually. Therefore, both the number of messages and the number of steps is increased by a factor of $n$.

\subsection{Measurements}
The Flood and Echo Net is implemented using PyTorch Lightning and PyG, the code will be made publicly available upon acceptance. The flood and echo are implemented in such a way, that they make use of the GPU operations provided by PyTorch Geometric by masking out the non-relevant messages. This precomputation implemented through message-passing on the GPU as well.
In Table \ref{app:runtime} we measure the execution of the forward pass of all models on the PrefixSum task for graphs of size $10, 20$, $50$, and $100$. Each run consists of a 1000 graphs for which we report the mean execution time per graph and the standard deviation.

Note that we take the exact same setup as in the PrefixSum task. Therefore, GIN always executes 5 layers and its runtime is not really impacted on larger graph. The RecGNN baseline performs $1.2n$ rounds of message-passing, where $n$ denotes the graph size. As the graphs grow larger, the runtime increases roughly linear. A similar behaviour can be seen in the \textit{random} and \textit{fixed} variations of the Flood and Echo Net. Note, that they execute two phases, each consisting of a flooding and echoing part. Therefore, there are about $4n$ steps of message-passing. Together with the precomputation of the distances for appropriate masking, this can account for the relative difference in performance. The \textit{all} variation of Flood and Echo performs $n$ single executions in a sequential order. It might be possible to at least partially parallelize these executions. However, as the number of different runs scales with the number of nodes, we believe that the fixed and random variants of the Flood and Echo Net are more suited for the study of extrapolation.

Note that in this specific experiment, the diameter of the graph is $n$. Due to the way the mechanism couples the number of iterations to the graph diameter, this is the worst case scenario. Therefore, we expect the performance ratio compared to RecGNN (which scales the number of iterations to the graph size) to be upper bounded by our measurements. While the current implementation is a bit slower compared to the standard MPNNS, due to the GPU support, the performance is still reasonable and practical for further research. Further, recall that the achieved performance of the models drastically differ. Moreover, while this is not yet the case for the current implementation, future implementations could leverage that the set of simultaneously active nodes is much smaller than the graph itself. This could drastically improve the overall usage of the GPU memory and open up further applications.

\begin{table*}
\centering
\caption{Runtime measurements performed on the PrefixSum task on $1000$ graphs per graph size. We report the mean time per graph in ms and the corresponding standard deviation. All measurements were performed on a NVIDIA GeForce RTX 3090.}
\resizebox{0.6\textwidth}{!}{
\begin{tabular}{@{}l*{5}{S[table-format=-3.4]}@{}}
\toprule
{Model} & \multicolumn{4}{c}{Time Measurement [ms]} \\
{}  & {n(10)} & {n(20)} & {n(50)} & {n(100)} \\
\midrule
GIN & {0.003 $\pm$ 0.023} & {0.003 $\pm$ 0.022} & {0.003 $\pm$ 0.027} & {0.003 $\pm$ 0.023} \\
RecGNN & {0.008 $\pm$ 0.023} & {0.015 $\pm$ 0.022} & {0.034 $\pm$ 0.028} & {0.066 $\pm$ 0.023} \\
\midrule
Flood and Echo \textit{all} & {0.304 $\pm$ 0.025} & {1.284 $\pm$ 0.031} & {7.995 $\pm$ 0.050} & {31.169 $\pm$ 0.168} \\
Flood and Echo \textit{random} & {0.031 $\pm$ 0.025} & {0.066 $\pm$ 0.033} & {0.160 $\pm$ 0.042} & {0.315 $\pm$ 0.066} \\
Flood and Echo \textit{fixed} & {0.040 $\pm$ 0.025} & {0.084 $\pm$ 0.030} & {0.212 $\pm$ 0.029} & {0.422 $\pm$ 0.029} \\

\bottomrule
\end{tabular}}
\label{tab:time_measurements}
\end{table*}  
\subsection{Standard Deviation of \textit{random} Variation}

\begin{table*}
\centering
\caption{Measurement of the standard deviation of the Flood and Echo \textit{random} variant. Each model performs 50 runs over 1000 graphs, we report the node and graph accuracy in percent as well as the minimum and maximum achieved accuracy for each model instance.}
\resizebox{0.6\textwidth}{!}{
\begin{tabular}{@{}l*{5}{S[table-format=-3.4]}@{}}
\toprule
{Model} & \multicolumn{4}{c}{\textsc{PrefixSum}} \\
{}  & {n(100)} & {min,max} & {g(100)} & {min,max} \\
\midrule
Model A & {98.78 $\pm$ 0.19} & {(98.28, 99.11)} & {96.43 $\pm$ 0.34} & {(95.70, 97.20)} \\
Model B & {100.00 $\pm$ 0.00} & {(100.00, 100.00)} & {100.00 $\pm$ 0.00} & {(100.00, 100.00)} \\
Model C & {100.00 $\pm$ 0.00} & {(100.00, 100.00)} & {100.00 $\pm$ 0.00} & {(100.00, 100.00)} \\
Model D & {91.37 $\pm$ 0.44} & {(90.40, 92.48)} & {74.97 $\pm$ 0.82} & {(73.50, 77.40)} \\
Model E & {100.00 $\pm$ 0.00} & {(100.00, 100.00)} & {100.00 $\pm$ 0.00} & {(100.00, 100.00)} \\

\bottomrule
\end{tabular}}
\label{tab:random}
\end{table*}  
By using the \textit{random} variant, we introduce a certain randomness in the computation, which could result in different outcomes depending on the chosen origin node. 

We measure the deviation of the random variant in the PrefixSum task. Each model performs 50 runs over 1000 graphs, we report the node and graph accuracy in percent as well as the minimum and maximum achieved accuracy for each model instance.
From the results in Table \ref{tab:random}, we can see that there are differences between the models, however, the variance due to the chosen origins within each model is quite small.

\section{Extrapolation}

In Table \ref{tab:path} we report the full results for the Path-Finding task and in Table \ref{tab:distance} for the Distance task.
\begin{table*}[h!]
\centering
\caption{Extrapolation on the Distance task. All models are trained with graphs of size 10 and then tested on larger graphs. The Flood and Echo models are able to generalize well to graphs $100$ times the sizes encountered during training. We report both the node accuracy with $n()$ and the graph accuracy with $g()$. }
\resizebox{1.0\textwidth}{!}{
\begin{tabular}{@{}l*{8}{S[table-format=-3.4]}@{}}
\toprule
{Model} & {\textsc{Messages}} & \multicolumn{6}{c}{\textsc{Distance}} \\
{} & {} & {n(10)} & {g(10)} & {n(100)} & {g(100)}  & {n(1000)} & {g(1000)}\\
\midrule  
GIN & {$\oo(Lm)$} &{0.99 $\pm$ 0.01 } & {0.92 $\pm$ 0.06 }& {0.70 $\pm$ 0.05 } & {0.00 $\pm$ 0.00 }& {0.53 $\pm$ 0.01 } & {0.00 $\pm$ 0.00 }\\
PGN & {$\oo(nm)$} & {1.00 $\pm$ 0.00 } & {1.00 $\pm$ 0.00 }& {0.77 $\pm$ 0.03 } & {0.00 $\pm$ 0.00 }& {0.50 $\pm$ 0.00 } & {0.00 $\pm$ 0.00 }\\
RecGNN & {$\oo(nm)$} & {1.00 $\pm$ 0.00 } & {1.00 $\pm$ 0.00 }& {0.95 $\pm$ 0.04 } & {0.45 $\pm$ 0.33 }& {0.78 $\pm$ 0.13 } & {0.00 $\pm$ 0.00 }\\
%\midrule
%FloodEcho-all& {$\oo(nm)$} \textcolor{red}{probably cut because not enough time}\\
\midrule
Flood and Echo \textit{random} & {$\oo(m)$} & {1.00 $\pm$ 0.00 } & {1.00 $\pm$ 0.00 }& {0.82 $\pm$ 0.01 } & {0.01 $\pm$ 0.00 }& {0.58 $\pm$ 0.01 } & {0.00 $\pm$ 0.00 }\\
Flood and Echo \textit{fixed}& {$\oo(m)$} &  {1.00 $\pm$ 0.00 } & {1.00 $\pm$ 0.00 }& {1.00 $\pm$ 0.00 } & {1.00 $\pm$ 0.00 }& {1.00 $\pm$ 0.00 } & {1.00 $\pm$ 0.00 }\\
\bottomrule
\end{tabular}}
\label{tab:distance}
\end{table*}

\begin{table*}[h!]
\centering
\caption{Extrapolation on the Path-Finding task. All models are trained with graphs of size 10 and then tested on larger graphs. The Flood and Echo models are able to generalize well to graphs $100$ times the sizes encountered during training. We report both the node accuracy with $n()$ and the graph accuracy with $g()$. }
\resizebox{1.0\textwidth}{!}{
\begin{tabular}{@{}l*{8}{S[table-format=-3.4]}@{}}
\toprule
{Model} & {\textsc{Messages}} & \multicolumn{6}{c}{\textsc{Path-Finding}} \\
{} & {} & {n(10)} & {g(10)} & {n(100)} & {g(100)}  & {n(1000)} & {g(1000)}\\
\midrule  
GIN & {$\oo(Lm)$} & {0.97 $\pm$ 0.01 } & {0.77 $\pm$ 0.08 }& {0.91 $\pm$ 0.01 } & {0.04 $\pm$ 0.06 }& {0.95 $\pm$ 0.01 } & {0.00 $\pm$ 0.01 }\\
PGN & {$\oo(nm)$} & {0.99 $\pm$ 0.01 } & {0.91 $\pm$ 0.05 }& {0.89 $\pm$ 0.01 } & {0.01 $\pm$ 0.02 }& {0.96 $\pm$ 0.00 } & {0.00 $\pm$ 0.00 }\\
RecGNN & {$\oo(nm)$} & {1.00 $\pm$ 0.00 } & {1.00 $\pm$ 0.00 }& {0.99 $\pm$ 0.02 } & {0.93 $\pm$ 0.15 }& {0.99 $\pm$ 0.01 } & {0.79 $\pm$ 0.37 }\\
%\midrule
%FloodEcho-all& {$\oo(nm)$} \textcolor{red}{probably cut because not enough time}\\
\midrule
Flood and Echo \textit{random}& {$\oo(m)$} & {1.00 $\pm$ 0.00 } & {1.00 $\pm$ 0.00 }& {0.97 $\pm$ 0.04 } & {0.77 $\pm$ 0.30 }& {0.98 $\pm$ 0.02 } & {0.48 $\pm$ 0.38 }\\
Flood and Echo \textit{fixed}& {$\oo(m)$} & {1.00 $\pm$ 0.00 } & {1.00 $\pm$ 0.00 }& {1.00 $\pm$ 0.00 } & {0.99 $\pm$ 0.02 }& {1.00 $\pm$ 0.00 } & {0.89 $\pm$ 0.13 }\\
\bottomrule
\end{tabular}}
\label{tab:path}
\end{table*}  

\section{SALSA}
\label{app:salsa}

We follow the training setup from \citet{minder2023salsaclrs}. If not specified otherwise, we run a single phase of the Flood and Echo Net using batchsize 8, max aggregation, the AdamW optimizer with an initial learning rate around 0.0004 while also reducing the learning rate by a factor of 0.1 if the validation loss does not decrease for 10 epochs. We employ an early stop if the validation loss does not decrease for 25 epochs and run the training for at most 100 epochs. All reported mean accuracies are taken across 5 model run on a NVIDIA GeForce RTX 3090.

The full results for all tasks on all graph distributions is depicted in Table \ref{tab:full_node_accuracy} for node accuracy and in Table \ref{tab:full_graph_accuracy} for graph accuracy. Further in Tables \ref{tab:mis-rounds-node},\ref{tab:mis-rounds-graph},\ref{tab:dijkstra-rounds-node} and \ref{tab:dijkstra-rounds-graph} we report the exact figures for the performance on MIS and Dijkstra if the number of rounds is increased.

\begin{table}[h!]
    \centering
    \caption{We test the Flood and Echo Net on the SALSA-CLRS benchmark across six graph based algorithmic tasks. All models are trained on graphs of size 16 and then tested on larger graphs. We report the node accuracy on Erd\H{o}s--R\'enyi graphs of different sizes. All numbers are taken across 5 runs.}
    
\resizebox{1.0\linewidth}{!}{

\begin{tabular}{@{}l*{17}{S[table-format=-3.4]}@{}}
            \toprule
            {Task} & {Model} & \multicolumn{5}{c}{\textsc{ER}} & \multicolumn{5}{c}{\textsc{WS}} & \multicolumn{5}{c}{\textsc{Delaunay}}\\{16} & {80} & {160} & {800} & {1600}& {16} & {80} & {160} & {800} & {1600}& {16} & {80} & {160} & {800} & {1600}\\
\midrule
 BFS & {Flood and Echo} & {{100.0 \text{\tiny $\pm$ 0.0} }} & {{100.0 \text{\tiny $\pm$ 0.0} }} & {{100.0 \text{\tiny $\pm$ 0.0} }} & {{99.7 \text{\tiny $\pm$ 0.1} }} & {{99.6 \text{\tiny $\pm$ 0.2} }} & {{100.0 \text{\tiny $\pm$ 0.0} }} & {{98.8 \text{\tiny $\pm$ 0.7} }} & {{96.6 \text{\tiny $\pm$ 2.0} }} & {{87.2 \text{\tiny $\pm$ 4.6} }} & {{86.8 \text{\tiny $\pm$ 1.4} }} & {{100.0 \text{\tiny $\pm$ 0.0} }} & {{99.6 \text{\tiny $\pm$ 0.3} }} & {{95.9 \text{\tiny $\pm$ 3.0} }} & {{62.1 \text{\tiny $\pm$ 6.7} }} & {{50.4 \text{\tiny $\pm$ 4.4} }} \\
 & {GIN(E)} & {99.4 \text{\tiny $\pm$ 0.8}} & {84.3 \text{\tiny $\pm$ 13.9}} & {57.5 \text{\tiny $\pm$ 15.3}} & {2.2 \text{\tiny $\pm$ 4.1}} & {0.1 \text{\tiny $\pm$ 0.2}} & {98.0 \text{\tiny $\pm$ 4.2}} & {5.7 \text{\tiny $\pm$ 8.7}} & {0.2 \text{\tiny $\pm$ 0.5}} & {0.0 \text{\tiny $\pm$ 0.0}} & {0.0 \text{\tiny $\pm$ 0.0}} & {99.3 \text{\tiny $\pm$ 1.0}} & {25.1 \text{\tiny $\pm$ 28.6}} & {0.7 \text{\tiny $\pm$ 1.4}} & {0.0 \text{\tiny $\pm$ 0.0}} & {0.0 \text{\tiny $\pm$ 0.0}} \\
 & {PGN} & {100.0 \text{\tiny $\pm$ 0.0}} & {88.7 \text{\tiny $\pm$ 5.9}} & {54.9 \text{\tiny $\pm$ 21.5}} & {0.2 \text{\tiny $\pm$ 0.1}} & {0.0 \text{\tiny $\pm$ 0.0}} & {100.0 \text{\tiny $\pm$ 0.0}} & {13.1 \text{\tiny $\pm$ 3.3}} & {0.1 \text{\tiny $\pm$ 0.1}} & {0.0 \text{\tiny $\pm$ 0.0}} & {0.0 \text{\tiny $\pm$ 0.0}} & {100.0 \text{\tiny $\pm$ 0.0}} & {35.1 \text{\tiny $\pm$ 8.3}} & {0.3 \text{\tiny $\pm$ 0.4}} & {0.0 \text{\tiny $\pm$ 0.0}} & {0.0 \text{\tiny $\pm$ 0.0}} \\
\bottomrule
\end{tabular}
}
    \label{tab:full_node_accuracy}
\end{table}

\begin{table}[h!]
    \centering
    \caption{We test the Flood and Echo Net on the SALSA-CLRS benchmark across six graph based algorithmic tasks. All models are trained on graphs of size 16 and then tested on larger graphs. We report the graph accuracy on Erd\H{o}s--R\'enyi graphs of different sizes. All numbers are taken across 5 runs.}
    
\resizebox{1.0\linewidth}{!}{

        \begin{tabular}{@{}l*{17}{S[table-format=-3.4]}@{}}
            \toprule
            {Task} & {Model} & \multicolumn{5}{c}{\textsc{ER}} & \multicolumn{5}{c}{\textsc{WS}} & \multicolumn{5}{c}{\textsc{Delaunay}}\\& & {16} & {80} & {160} & {800} & {1600}& {16} & {80} & {160} & {800} & {1600}& {16} & {80} & {160} & {800} & {1600}\\
\midrule
 BFS & {Flood and Echo} & {{100.0 \text{\tiny $\pm$ 0.0} }} & {{98.9 \text{\tiny $\pm$ 1.2} }} & {{95.1 \text{\tiny $\pm$ 3.4} }} & {{26.1 \text{\tiny $\pm$ 16.2} }} & {{6.1 \text{\tiny $\pm$ 6.2} }} & {{100.0 \text{\tiny $\pm$ 0.0} }} & {{52.7 \text{\tiny $\pm$ 20.3} }} & {{14.8 \text{\tiny $\pm$ 20.2} }} & {{0.0 \text{\tiny $\pm$ 0.0} }} & {{0.0 \text{\tiny $\pm$ 0.0} }} & {{100.0 \text{\tiny $\pm$ 0.0} }} & {{77.2 \text{\tiny $\pm$ 13.8} }} & {{19.7 \text{\tiny $\pm$ 22.2} }} & {{0.0 \text{\tiny $\pm$ 0.0} }} & {{0.0 \text{\tiny $\pm$ 0.0} }} \\
 & {GIN(E)} & {99.4 \text{\tiny $\pm$ 0.8}} & {84.3 \text{\tiny $\pm$ 13.9}} & {57.5 \text{\tiny $\pm$ 15.3}} & {2.2 \text{\tiny $\pm$ 4.1}} & {0.1 \text{\tiny $\pm$ 0.2}} & {98.0 \text{\tiny $\pm$ 4.2}} & {5.7 \text{\tiny $\pm$ 8.7}} & {0.2 \text{\tiny $\pm$ 0.5}} & {0.0 \text{\tiny $\pm$ 0.0}} & {0.0 \text{\tiny $\pm$ 0.0}} & {99.3 \text{\tiny $\pm$ 1.0}} & {25.1 \text{\tiny $\pm$ 28.6}} & {0.7 \text{\tiny $\pm$ 1.4}} & {0.0 \text{\tiny $\pm$ 0.0}} & {0.0 \text{\tiny $\pm$ 0.0}} \\
 & {PGN} & {100.0 \text{\tiny $\pm$ 0.0}} & {88.7 \text{\tiny $\pm$ 5.9}} & {54.9 \text{\tiny $\pm$ 21.5}} & {0.2 \text{\tiny $\pm$ 0.1}} & {0.0 \text{\tiny $\pm$ 0.0}} & {100.0 \text{\tiny $\pm$ 0.0}} & {13.1 \text{\tiny $\pm$ 3.3}} & {0.1 \text{\tiny $\pm$ 0.1}} & {0.0 \text{\tiny $\pm$ 0.0}} & {0.0 \text{\tiny $\pm$ 0.0}} & {100.0 \text{\tiny $\pm$ 0.0}} & {35.1 \text{\tiny $\pm$ 8.3}} & {0.3 \text{\tiny $\pm$ 0.4}} & {0.0 \text{\tiny $\pm$ 0.0}} & {0.0 \text{\tiny $\pm$ 0.0}} \\
\bottomrule
\end{tabular}
}
    \label{tab:full_graph_accuracy}
\end{table}

\begin{table}[h!]
    \centering
    \caption{Results for the Flood and Echo Net on the MIS task when the number of rounds is increased. We report node accuracy, SALSA-CLRS indicates that the number of phases matches the length of the algorithm trajectory.}
    \resizebox{1.0\linewidth}{!}{

        \begin{tabular}{@{}l*{17}{S[table-format=-3.4]}@{}}
            \toprule
            {Model} & {Task} & \multicolumn{5}{c}{\textsc{ER}} & \multicolumn{5}{c}{\textsc{WS}} & \multicolumn{5}{c}{\textsc{Delaunay}}\\& & {16} & {80} & {160} & {800} & {1600}& {16} & {80} & {160} & {800} & {1600}& {16} & {80} & {160} & {800} & {1600}\\
            \midrule
Flood and Echo Net - SALSA-CLRS & MIS & {99.7 \text{\tiny $\pm$ 0.1} } & {99.5 \text{\tiny $\pm$ 0.1} } & {99.4 \text{\tiny $\pm$ 0.2} } & {98.9 \text{\tiny $\pm$ 0.3} } & {99.2 \text{\tiny $\pm$ 0.2} } & {99.8 \text{\tiny $\pm$ 0.1} } & {99.8 \text{\tiny $\pm$ 0.0} } & {99.7 \text{\tiny $\pm$ 0.1} } & {99.7 \text{\tiny $\pm$ 0.1} } & {99.7 \text{\tiny $\pm$ 0.1} } & {99.8 \text{\tiny $\pm$ 0.0} } & {99.8 \text{\tiny $\pm$ 0.1} } & {99.7 \text{\tiny $\pm$ 0.0} } & {99.7 \text{\tiny $\pm$ 0.1} } & {99.7 \text{\tiny $\pm$ 0.1} } \\
Flood and Echo Net - 1 & MIS & {91.5 \text{\tiny $\pm$ 0.1} } & {87.5 \text{\tiny $\pm$ 0.2} } & {87.8 \text{\tiny $\pm$ 0.1} } & {88.3 \text{\tiny $\pm$ 0.3} } & {87.4 \text{\tiny $\pm$ 0.2} } & {92.7 \text{\tiny $\pm$ 0.2} } & {91.9 \text{\tiny $\pm$ 0.1} } & {91.6 \text{\tiny $\pm$ 0.2} } & {92.5 \text{\tiny $\pm$ 0.3} } & {92.1 \text{\tiny $\pm$ 0.2} } & {92.9 \text{\tiny $\pm$ 0.1} } & {91.8 \text{\tiny $\pm$ 0.1} } & {91.5 \text{\tiny $\pm$ 0.2} } & {91.5 \text{\tiny $\pm$ 0.2} } & {91.4 \text{\tiny $\pm$ 0.2} } \\
Flood and Echo Net - 2 & MIS & {93.1 \text{\tiny $\pm$ 0.1} } & {89.4 \text{\tiny $\pm$ 0.1} } & {90.1 \text{\tiny $\pm$ 0.0} } & {89.5 \text{\tiny $\pm$ 0.1} } & {89.3 \text{\tiny $\pm$ 0.1} } & {93.8 \text{\tiny $\pm$ 0.1} } & {93.8 \text{\tiny $\pm$ 0.1} } & {93.5 \text{\tiny $\pm$ 0.0} } & {94.2 \text{\tiny $\pm$ 0.0} } & {93.9 \text{\tiny $\pm$ 0.0} } & {94.1 \text{\tiny $\pm$ 0.1} } & {93.5 \text{\tiny $\pm$ 0.0} } & {93.2 \text{\tiny $\pm$ 0.0} } & {93.2 \text{\tiny $\pm$ 0.0} } & {93.2 \text{\tiny $\pm$ 0.0} } \\
\bottomrule
\end{tabular}
}
    \label{tab:mis-rounds-node}
\end{table}
\begin{table}[h!]
    \centering
    \caption{Results for the Flood and Echo Net on the MIS task when the number of rounds is increased. We report graph accuracy, SALSA-CLRS indicates that the number of phases matches the length of the algorithm trajectory.}
    \resizebox{1.0\linewidth}{!}{

        \begin{tabular}{@{}l*{17}{S[table-format=-3.4]}@{}}
            \toprule
            {Model} & {Task} & \multicolumn{5}{c}{\textsc{ER}} & \multicolumn{5}{c}{\textsc{WS}} & \multicolumn{5}{c}{\textsc{Delaunay}}\\& & {16} & {80} & {160} & {800} & {1600}& {16} & {80} & {160} & {800} & {1600}& {16} & {80} & {160} & {800} & {1600}\\
            \midrule
Flood and Echo Net - SALSA-CLRS & MIS & {98.3 \text{\tiny $\pm$ 0.6} } & {90.0 \text{\tiny $\pm$ 2.0} } & {81.0 \text{\tiny $\pm$ 4.4} } & {24.3 \text{\tiny $\pm$ 10.4} } & {10.9 \text{\tiny $\pm$ 6.6} } & {98.3 \text{\tiny $\pm$ 0.7} } & {95.1 \text{\tiny $\pm$ 1.6} } & {85.6 \text{\tiny $\pm$ 1.7} } & {48.0 \text{\tiny $\pm$ 7.4} } & {24.1 \text{\tiny $\pm$ 7.3} } & {98.8 \text{\tiny $\pm$ 0.3} } & {93.7 \text{\tiny $\pm$ 1.2} } & {87.4 \text{\tiny $\pm$ 1.2} } & {48.6 \text{\tiny $\pm$ 7.0} } & {24.3 \text{\tiny $\pm$ 4.7} } \\
Flood and Echo Net - 1 & MIS & {40.4 \text{\tiny $\pm$ 2.1} } & {0.1 \text{\tiny $\pm$ 0.1} } & {0.0 \text{\tiny $\pm$ 0.0} } & {0.0 \text{\tiny $\pm$ 0.0} } & {0.0 \text{\tiny $\pm$ 0.0} } & {40.3 \text{\tiny $\pm$ 2.4} } & {1.0 \text{\tiny $\pm$ 0.3} } & {0.0 \text{\tiny $\pm$ 0.0} } & {0.0 \text{\tiny $\pm$ 0.0} } & {0.0 \text{\tiny $\pm$ 0.0} } & {44.1 \text{\tiny $\pm$ 1.2} } & {1.2 \text{\tiny $\pm$ 0.1} } & {0.0 \text{\tiny $\pm$ 0.0} } & {0.0 \text{\tiny $\pm$ 0.0} } & {0.0 \text{\tiny $\pm$ 0.0} } \\
Flood and Echo Net - 2 & MIS & {49.1 \text{\tiny $\pm$ 2.0} } & {0.3 \text{\tiny $\pm$ 0.2} } & {0.0 \text{\tiny $\pm$ 0.0} } & {0.0 \text{\tiny $\pm$ 0.0} } & {0.0 \text{\tiny $\pm$ 0.0} } & {46.3 \text{\tiny $\pm$ 1.6} } & {3.1 \text{\tiny $\pm$ 0.6} } & {0.0 \text{\tiny $\pm$ 0.0} } & {0.0 \text{\tiny $\pm$ 0.0} } & {0.0 \text{\tiny $\pm$ 0.0} } & {49.1 \text{\tiny $\pm$ 1.7} } & {2.2 \text{\tiny $\pm$ 0.5} } & {0.1 \text{\tiny $\pm$ 0.1} } & {0.0 \text{\tiny $\pm$ 0.0} } & {0.0 \text{\tiny $\pm$ 0.0} } \\
\bottomrule
\end{tabular}
}
    \label{tab:mis-rounds-graph}
\end{table}
\begin{table}[h!]
    \centering
    \caption{Results for the Flood and Echo Net on the Dijkstra task when the number of rounds is increased. We report node accuracy, SALSA-CLRS indicates that the number of phases matches the length of the algorithm trajectory.}
    
\resizebox{1.0\linewidth}{!}{

        \begin{tabular}{@{}l*{17}{S[table-format=-3.4]}@{}}
            \toprule
            {Model} & {Task} & \multicolumn{5}{c}{\textsc{ER}} & \multicolumn{5}{c}{\textsc{WS}} & \multicolumn{5}{c}{\textsc{Delaunay}}\\& & {16} & {80} & {160} & {800} & {1600}& {16} & {80} & {160} & {800} & {1600}& {16} & {80} & {160} & {800} & {1600}\\
            \midrule
Flood and Echo Net - SALSA-CLRS & MIS & {99.5 \text{\tiny $\pm$ nan} } & {97.0 \text{\tiny $\pm$ nan} } & {93.7 \text{\tiny $\pm$ nan} } & {86.8 \text{\tiny $\pm$ nan} } & {83.4 \text{\tiny $\pm$ nan} } & {98.6 \text{\tiny $\pm$ nan} } & {96.9 \text{\tiny $\pm$ nan} } & {95.0 \text{\tiny $\pm$ nan} } & {85.2 \text{\tiny $\pm$ nan} } & {79.3 \text{\tiny $\pm$ nan} } & {99.3 \text{\tiny $\pm$ nan} } & {96.5 \text{\tiny $\pm$ nan} } & {92.6 \text{\tiny $\pm$ nan} } & {72.4 \text{\tiny $\pm$ nan} } & {61.4 \text{\tiny $\pm$ nan} } \\
Flood and Echo Net - 1 & MIS & {98.3 \text{\tiny $\pm$ 0.3} } & {89.0 \text{\tiny $\pm$ 0.9} } & {80.2 \text{\tiny $\pm$ 1.1} } & {64.8 \text{\tiny $\pm$ 1.0} } & {59.5 \text{\tiny $\pm$ 0.9} } & {96.2 \text{\tiny $\pm$ 0.6} } & {91.0 \text{\tiny $\pm$ 0.7} } & {88.1 \text{\tiny $\pm$ 0.7} } & {76.5 \text{\tiny $\pm$ 1.4} } & {71.7 \text{\tiny $\pm$ 2.0} } & {97.8 \text{\tiny $\pm$ 0.3} } & {90.4 \text{\tiny $\pm$ 1.3} } & {83.2 \text{\tiny $\pm$ 1.3} } & {62.6 \text{\tiny $\pm$ 2.5} } & {56.8 \text{\tiny $\pm$ 3.6} } \\
Flood and Echo Net - 2 & MIS & {99.0 \text{\tiny $\pm$ 0.1} } & {93.4 \text{\tiny $\pm$ 0.2} } & {87.3 \text{\tiny $\pm$ 0.5} } & {74.4 \text{\tiny $\pm$ 0.6} } & {69.5 \text{\tiny $\pm$ 0.5} } & {97.6 \text{\tiny $\pm$ 0.3} } & {93.8 \text{\tiny $\pm$ 0.4} } & {91.4 \text{\tiny $\pm$ 0.4} } & {80.6 \text{\tiny $\pm$ 1.4} } & {76.4 \text{\tiny $\pm$ 1.5} } & {98.7 \text{\tiny $\pm$ 0.2} } & {93.4 \text{\tiny $\pm$ 0.4} } & {88.2 \text{\tiny $\pm$ 0.8} } & {72.2 \text{\tiny $\pm$ 1.0} } & {66.0 \text{\tiny $\pm$ 1.1} } \\
\bottomrule
\end{tabular}
}
    \label{tab:dijkstra-rounds-node}
\end{table}
\begin{table}[h!]
    \centering
    \caption{Results for the Flood and Echo Net on the Dijkstra task when the number of rounds is increased. We report graph accuracy, SALSA-CLRS indicates that the number of phases matches the length of the algorithm trajectory.}
    \resizebox{1.0\linewidth}{!}{

        \begin{tabular}{@{}l*{17}{S[table-format=-3.4]}@{}}
            \toprule
            {Model} & {Task} & \multicolumn{5}{c}{\textsc{ER}} & \multicolumn{5}{c}{\textsc{WS}} & \multicolumn{5}{c}{\textsc{Delaunay}}\\& & {16} & {80} & {160} & {800} & {1600}& {16} & {80} & {160} & {800} & {1600}& {16} & {80} & {160} & {800} & {1600}\\
            \midrule
Flood and Echo Net - SALSA-CLRS & MIS & {93.1 \text{\tiny $\pm$ nan} } & {15.5 \text{\tiny $\pm$ nan} } & {0.7 \text{\tiny $\pm$ nan} } & {0.0 \text{\tiny $\pm$ nan} } & {0.0 \text{\tiny $\pm$ nan} } & {78.8 \text{\tiny $\pm$ nan} } & {13.9 \text{\tiny $\pm$ nan} } & {0.6 \text{\tiny $\pm$ nan} } & {0.0 \text{\tiny $\pm$ nan} } & {0.0 \text{\tiny $\pm$ nan} } & {89.9 \text{\tiny $\pm$ nan} } & {9.8 \text{\tiny $\pm$ nan} } & {0.0 \text{\tiny $\pm$ nan} } & {0.0 \text{\tiny $\pm$ nan} } & {0.0 \text{\tiny $\pm$ nan} } \\
Flood and Echo Net - 1 & MIS & {76.7 \text{\tiny $\pm$ 2.7} } & {1.0 \text{\tiny $\pm$ 0.4} } & {0.0 \text{\tiny $\pm$ 0.0} } & {0.0 \text{\tiny $\pm$ 0.0} } & {0.0 \text{\tiny $\pm$ 0.0} } & {56.8 \text{\tiny $\pm$ 4.0} } & {1.4 \text{\tiny $\pm$ 0.6} } & {0.0 \text{\tiny $\pm$ 0.0} } & {0.0 \text{\tiny $\pm$ 0.0} } & {0.0 \text{\tiny $\pm$ 0.0} } & {71.7 \text{\tiny $\pm$ 2.7} } & {0.2 \text{\tiny $\pm$ 0.2} } & {0.0 \text{\tiny $\pm$ 0.0} } & {0.0 \text{\tiny $\pm$ 0.0} } & {0.0 \text{\tiny $\pm$ 0.0} } \\
Flood and Echo Net - 2 & MIS & {86.3 \text{\tiny $\pm$ 1.1} } & {3.7 \text{\tiny $\pm$ 0.5} } & {0.0 \text{\tiny $\pm$ 0.1} } & {0.0 \text{\tiny $\pm$ 0.0} } & {0.0 \text{\tiny $\pm$ 0.0} } & {68.5 \text{\tiny $\pm$ 2.4} } & {3.5 \text{\tiny $\pm$ 0.7} } & {0.0 \text{\tiny $\pm$ 0.0} } & {0.0 \text{\tiny $\pm$ 0.0} } & {0.0 \text{\tiny $\pm$ 0.0} } & {81.7 \text{\tiny $\pm$ 2.3} } & {1.1 \text{\tiny $\pm$ 0.4} } & {0.0 \text{\tiny $\pm$ 0.0} } & {0.0 \text{\tiny $\pm$ 0.0} } & {0.0 \text{\tiny $\pm$ 0.0} } \\
\bottomrule
\end{tabular}
}
    \label{tab:dijkstra-rounds-graph}
\end{table}

\section{Datasets}
\subsection{Algorithmic Datasets}
\label{appendix:algorithmic_datasets}
For all the below tasks, we use train set, validation set, and test set sizes of {1024, 100, and 1000}, respectively. The sizes of the respective graphs in the train, validation, and test sets are {10, 20, and 100}. Performance on this test set demonstrates the model's ability to extrapolate to larger graph sizes. Note that many of the tasks only require the output modulo $2$. We reduce the problem to this specific setting so that all numbers involved in the computation stay within the same range, as otherwise, the values have to be interpreted almost in a symbolic way, which is very challenging for learning-based models.

\paragraph{PrefixSum Task}\citep{grötschla2022learning}
Each graph in this dataset is a path graph where each node has a random binary label with one marked vertex at one end, which indicates the starting point. The objective of this task is to predict whether the PrefixSum from the marked node to the node in consideration is divisible by 2. 

\begin{figure}
\centering
%\framebox[4.0in]{$\;$}
\includegraphics[width=\linewidth]{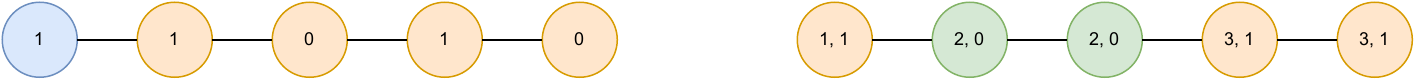}
\caption{Example graph from the PrefixSum task. The left graph represents the input graph with a binary value associated with each node and the blue node being the starting node. The right graph represents the ground truth solution, each node contains two values the cumulative sum and the desired result which is the cumulative sum modulo 2. }
\label{fig:prefix_sum}
\end{figure}

\paragraph{Distance Task}\citep{grötschla2022learning}
In this task every graph is a random graph of $n$ nodes with a source node being distinctly marked. The objective of this task is to predict for each node whether its distance to the source node is divisible by 2. 

\begin{figure}
\centering
\includegraphics[scale=0.40]{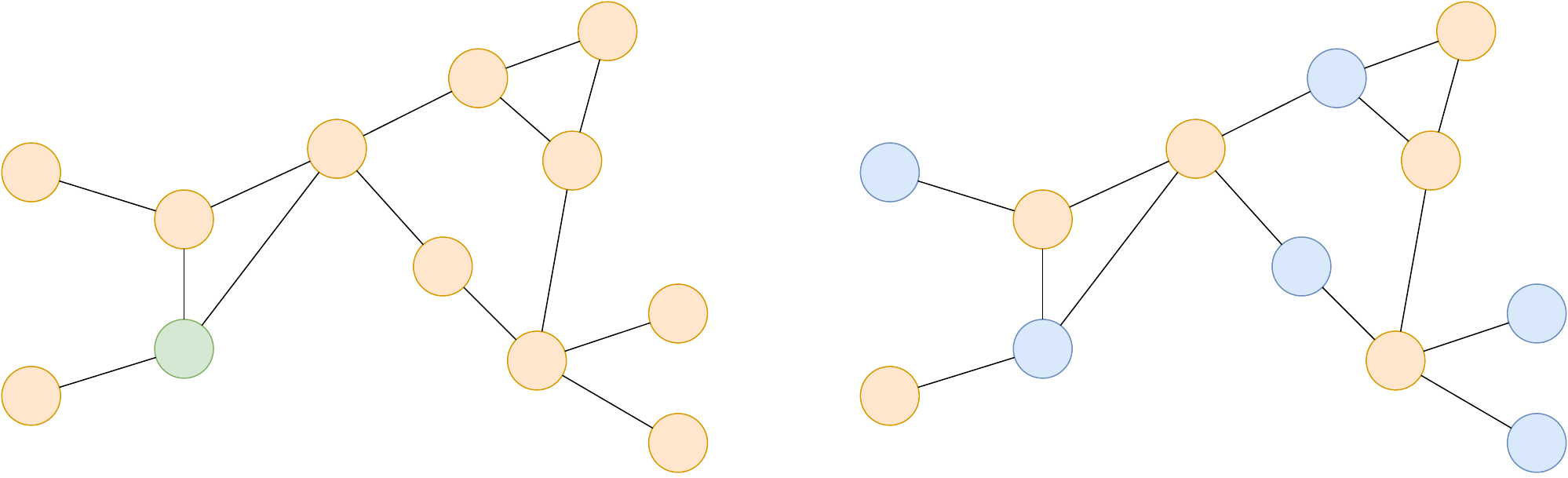}
\caption{Example graph from the distance task. The green node in the left graph (input graph) represents the source node, and the remaining nodes are unmarked. On the right graph (ground truth) all orange nodes are at an odd distance away from the source while the blue nodes are at an even distance away from the source.}
\label{fig:distance}
\end{figure}

\paragraph{Path Finding Task}\citep{grötschla2022learning}
In this task the dataset consists of random trees of $n$ nodes with two distinct vertices being marked separately. The objective of this task is to predict for each node whether it belongs to the shortest path between the 2 marked nodes. 

\begin{figure}
\centering
\includegraphics[width=\textwidth]{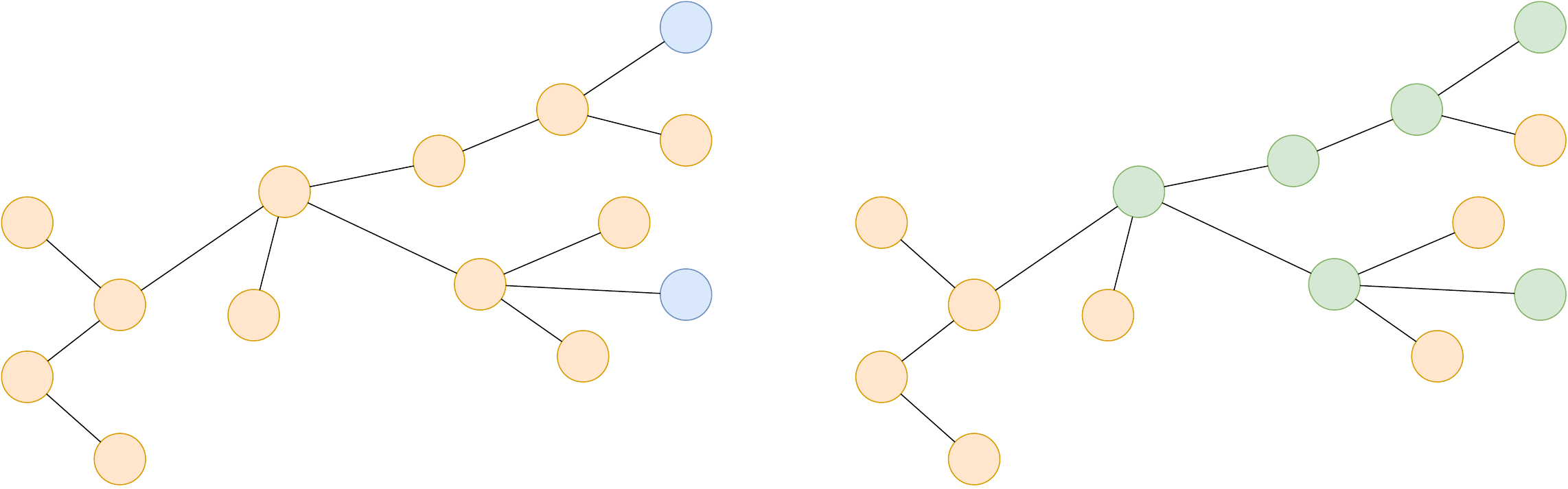}
\caption{Example graph from the pathfinding task. The left graph represents the input graph, where the blue nodes are the marked nodes. The right is the corresponding solution, where the path between the marked nodes is highlighted in green. }
\label{fig:path_finding}
\end{figure}

\subsection{Expressive Datasets}
\label{appendix:expressive_datasets}

\begin{figure}
\centering
\includegraphics[scale=0.20]{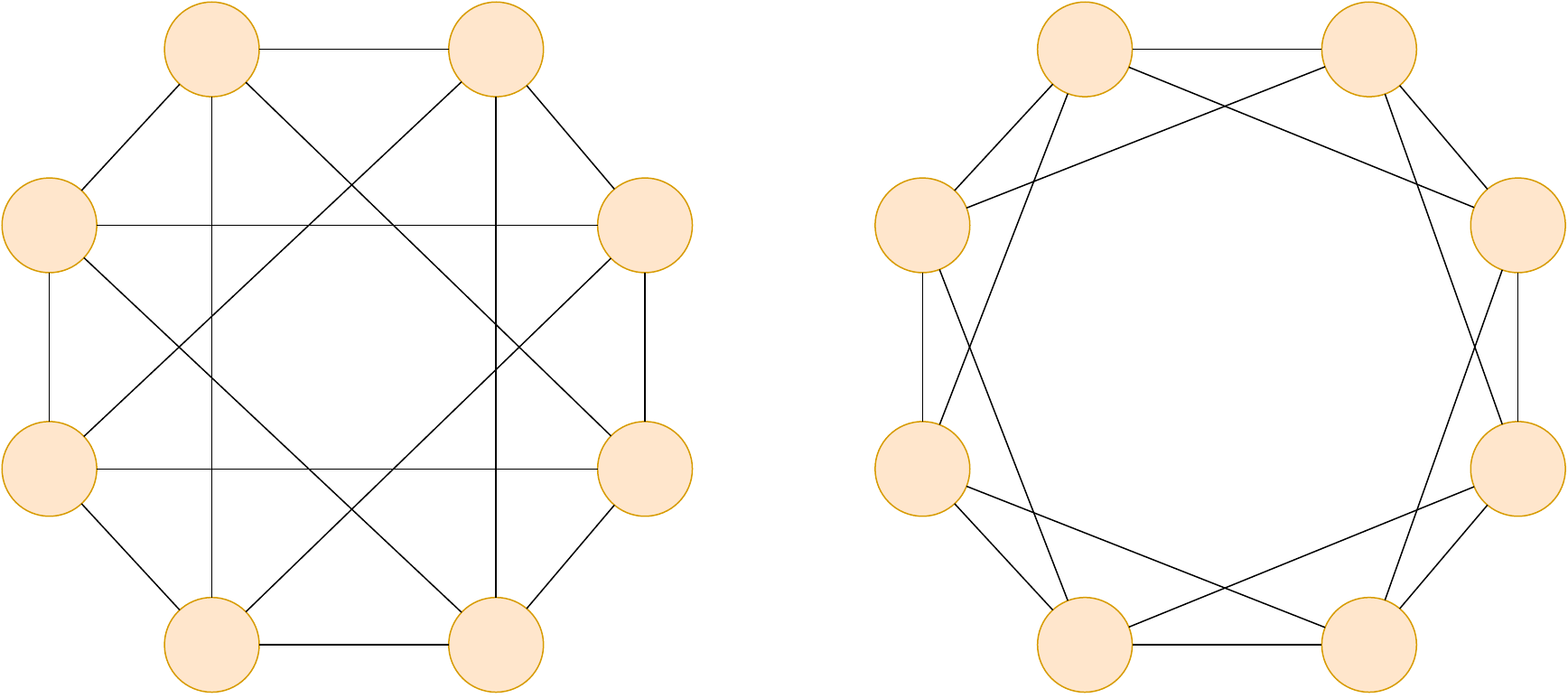}
\caption{Example graphs from the Skip Circles dataset, namely  $G_{n, 5}$ and $G_{n, 2}$ on the left and the right respectively.}
\label{fig:skip_circles}
\end{figure}

\paragraph{Skip Circles}\citep{chen2023equivalence}
This dataset consists of CSL(Circular Skip Link) graphs denoted by $G_{n, k}$, which is a graph of size $n$, numbered $0$ to $n-1$, where there exists an edge between node $i$ and node $j$ iff $|i - j| \equiv 1$ or $k$ (mod $n$). $G_{n, k}$ and $G_{n', k'}$ are only isomorphic when $n = n'$ and $k \equiv \pm k'$ (mod $n$). Here, the number of graphs in train, validation, and test are all 10.
We can see an example of this construction in Figure \ref{fig:skip_circles}.

We follow the setup of \citet{chen2023equivalence} where we fix $n = 41$ and set $k \in \{2, 3, 4, 5, 6, 9, 11, 12, 13, 16\}$. Each $G_{n, k}$ forms a separate isomorphism class, and the aim of the classification task is to classify the graph into its isomorphism class by the skip cycle length. Since 1-WL is unable to classify these graphs, we can see in table \ref{table:expressive} that the GIN model cannot get an accuracy better than random guessing ($10\%$). 
 
\begin{figure}%
    \centering
    \subfloat[\centering Limits 1]
 {{\includegraphics[width=6cm]{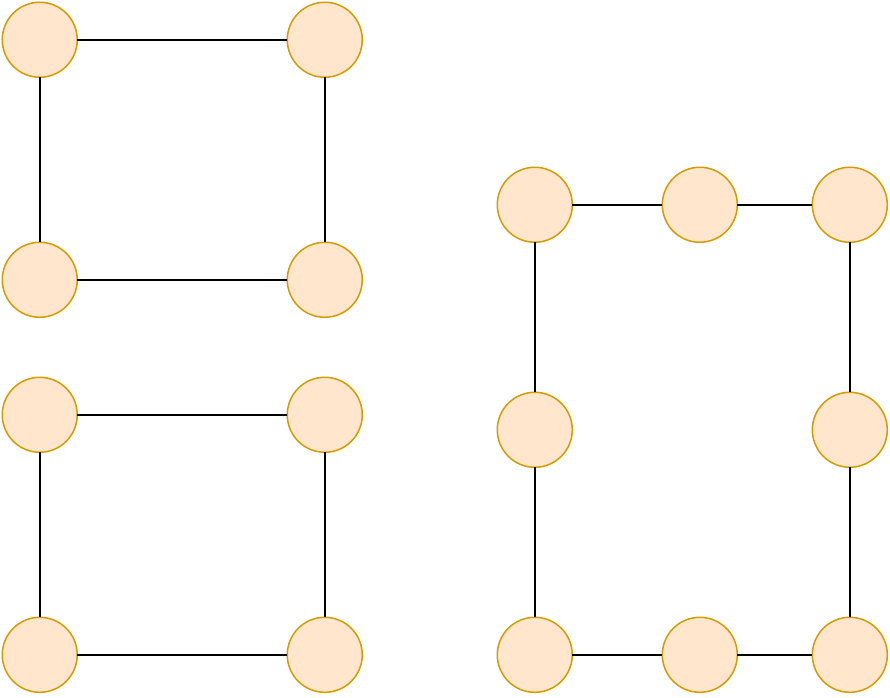} }}%
    \qquad
    \centering
    \subfloat[\centering Limits 2]
    {{\includegraphics[width=7cm]{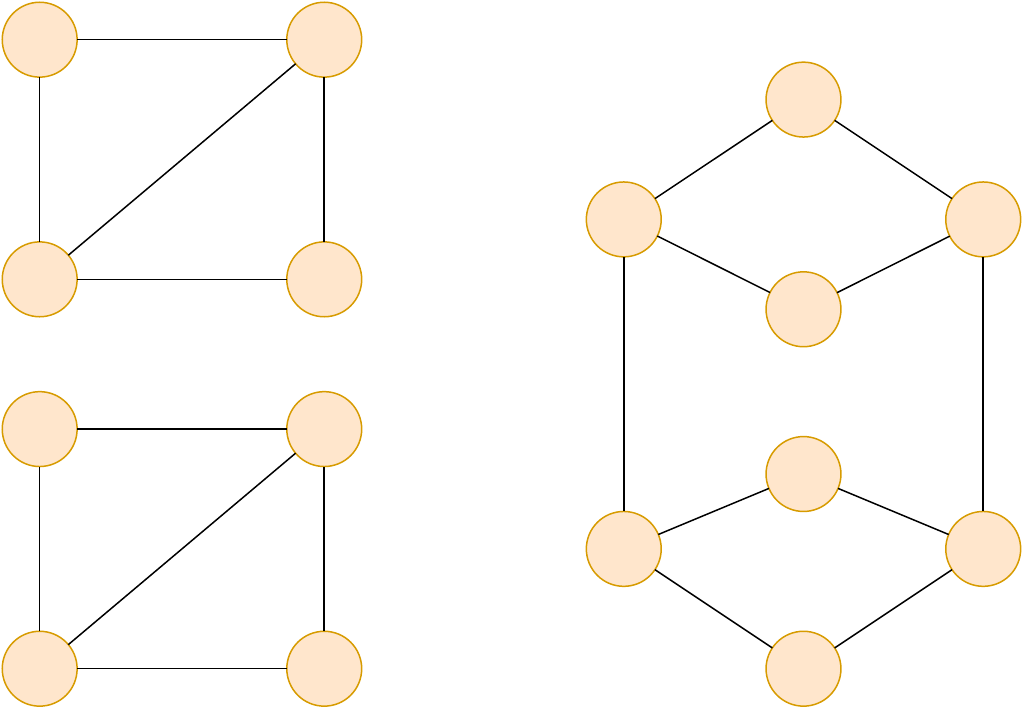} }}%
    \caption{Counter-examples which MPNNs cannot distinguish from \citet{garg2020generalization}, they cannot distinguish among the graphs in each example.}%
    \label{fig:limits}%
\end{figure}

\paragraph{Limits1 and Limits2}\citep{garg2020generalization}
This dataset consists of two graphs from \citet{garg2020generalization} that, despite having different girth, circumference, diameter, and total number of cycles, cannot be distinguished by 1-WL. For each example, the aim is to distinguish among the disjoint graphs on the left versus the larger component on the right. The specific constructions can be seen in Figure \ref{fig:limits}.

\begin{figure}
\centering
\includegraphics[scale=0.30]{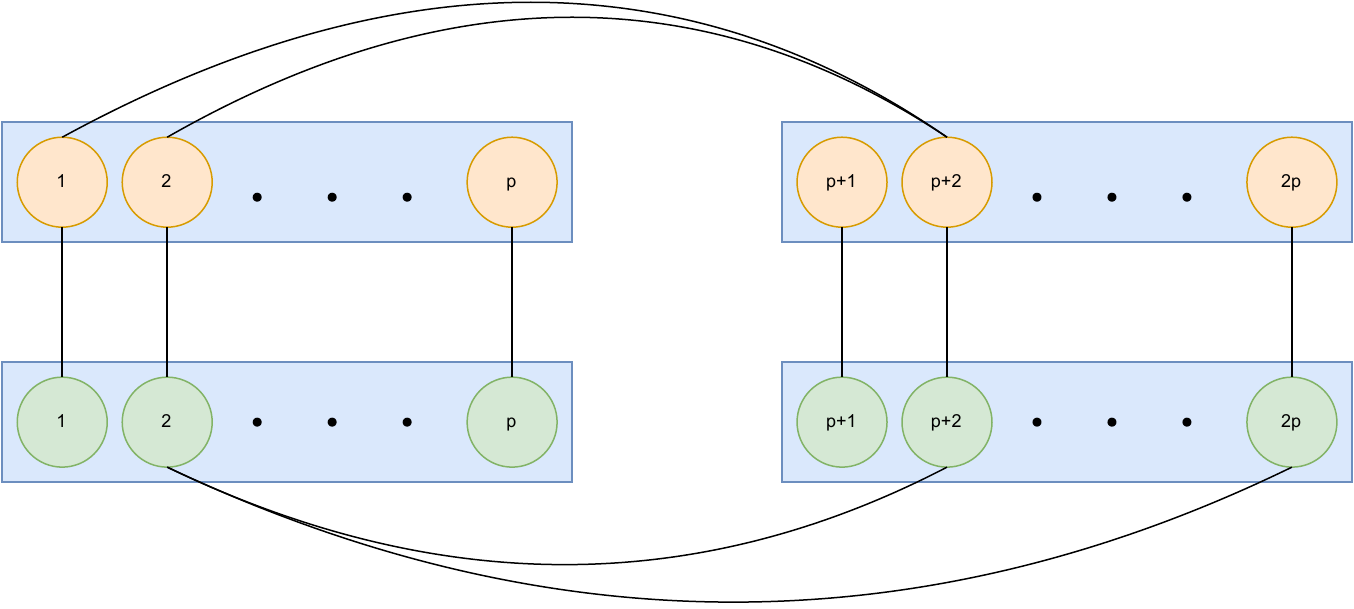}
\caption{Example construction of \citet{loukas2020graph}}, where k=4.
\label{fig:four_cycles}
\end{figure}

\paragraph{4-Cycles}\citep{loukas2020graph}
This dataset introduced by \citet{loukas2020graph} originates from a construction by \citet{korhonen2017deterministic} in which two players Alice and Bob each start with a complete bipartite graph of $p = \sqrt{q}$ nodes which are numbered from $1$ to $2p$ and a hidden binary key with size being $|p^2|$. The nodes from each graph with the same numbers are connected together. Each player then uses their respective binary keys to remove edges, each bipartite edge corresponding to a zero bit is removed and remaining edges are substituted by a path of length $k/2 - 1$, we use $k = 4$. The task is to determine if the resulting graph has a cycle of length $k$. In our implementation the number of train, validation and test graphs we consider are all 25. 
For a depiction of the construction refer to Figure \ref{fig:four_cycles}.

\begin{figure}
\centering
\includegraphics[scale=0.20]{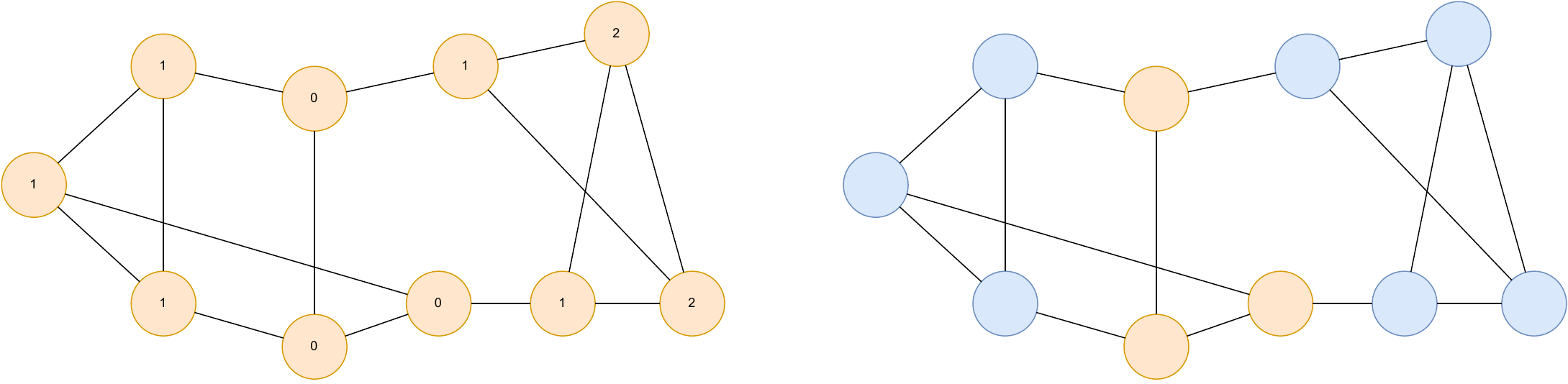}\caption{The graphs represent an instance from LLC and Triangles dataset respectively. For the LLC graph(left), each label denotes the ground truth for the graph while for the Triangles(right) graph, the blue nodes are ones which are a part of a triangle, while the orange nodes are not part of any triangle.}
\label{fig:LCC}
\end{figure}

\paragraph{LLC} \citep{sato2021random}
This dataset is comprised of random 3-regular graphs and the task is to determine for each node its local clustering coefficient \citep{watts1998collective} which informally is the number of triangles the vertex is part of. The training and test set are both comprised of a 1000 graphs. The graphs in the train set have 20 nodes, while the graphs in the test set have a 100 nodes testing extrapolation. An example graph from this dataset can be seen in Figure \ref{fig:LCC}.

\paragraph{Triangles} \citep{sato2021random}
This dataset akin to the previous contains random 3-regular graphs with the same train/test split and graph sizes. The task here is to classify each node as being part of a triangle or not. An example graph from this dataset can be seen in Figure \ref{fig:LCC}.

%%%%%%%%%%%%%%%%%%%%%%%%%%%%%%%%%%%%%%%%%%%%%%%%%%%%%%%%%%%%
\end{document}